\documentclass{article}

\PassOptionsToPackage{numbers, compress}{natbib}

\usepackage[preprint]{neurips_2024}

\usepackage[utf8]{inputenc} % allow utf-8 input
\usepackage[T1]{fontenc}    % use 8-bit T1 fonts
\usepackage{hyperref}       % hyperlinks
\usepackage{url}            % simple URL typesetting
\usepackage{booktabs}       % professional-quality tables
\usepackage{amsfonts}       % blackboard math symbols
\usepackage{nicefrac}       % compact symbols for 1/2, etc.
\usepackage{microtype}      % microtypography
\usepackage{xcolor}         % colors
\usepackage{graphicx}
\usepackage{hyperref}       % hyperlinks
\usepackage{url}            % simple URL typesetting
\usepackage{booktabs}       % professional-quality tables
\usepackage{amsfonts}       % blackboard math symbols
\usepackage{nicefrac}       % compact symbols for 1/2, etc.
\usepackage{microtype}      % microtypography
\usepackage{bm}
\usepackage{amsmath}
\usepackage{amssymb}
\usepackage{mathtools}
\usepackage{amsthm}
\usepackage{multirow}
\usepackage{colortbl}

\usepackage[linesnumbered,ruled,vlined]{algorithm2e}
\usepackage[ruled]{algorithm2e}
\usepackage{algpseudocode}  
\usepackage{amsmath} 
\usepackage{caption}
\usepackage{subfigure}
\usepackage{graphicx}
\usepackage{svg}
\usepackage{pifont}
\usepackage{authblk}

\usepackage{natbib}

\title{Preferred-Action-Optimized Diffusion Policies for Offline Reinforcement Learning} %superior action  Preferable

\author[1]{\textbf{Tianle Zhang}}
\author[1,2]{\textbf{Jiayi Guan}}
\author[1,3]{\textbf{Lin Zhao}}
\author[1]{\textbf{Yihang Li}}
\author[1]{\textbf{Dongjiang Li}}
\author[1]{\textbf{Zecui Zeng}}
\author[1]{\textbf{Lei Sun}}
\author[1]{\textbf{Yue Chen}}
\author[1]{\textbf{Xuelong Wei}}
\author[1]{\textbf{Lusong Li}}
\author[1]{\textbf{Xiaodong He}}
\affil[1]{JD Explore Academy, China}
\affil[2]{Tongji University, China}
\affil[3]{Beijing Institute of Technology, China}
\affil[ ]{\texttt{tianle-zhang@outlook.com, guanjiayi@tongji.edu.cn, zhaolins@foxmail.com, \{liyihang18, lidongjiang5, zengzecui1\}@jd.com, \{sunlei155, chenyue21, weixuelong1\}@jd.com, lilusong@gmail.com, xiaodong.he@jd.com}}

\begin{document}

\maketitle

\begin{abstract}

% 

 % Offline reinforcement learning (RL) aims to learn optimal policies from previously collected datasets. Recently, the offline RL works using diffusion models as policy models have shown great potential, due to the strong expressiveness of the diffusion models. However, they still adopt the weighted regression (WR) method for policy improvement. WR is confined to using sample actions to optimize policies and is sensitive to Q-value, resulting in limited or lower performance. To mitigate the problem, we propose a novel preferred-action-optimized diffusion policy for offline RL. In particular, an expressive conditional diffusion model is utilized to represent the diverse distribution of a behavior policy. Meanwhile, based on the diffusion model, preferred actions within the same behavior distribution are automatically generated through the critic function. Moreover, an anti-noise preference optimization is designed to achieve policy improvement by using the preferred actions, which can adapt to noise-preferred actions for stable training. Experimental results on D4RL benchmarking show that the proposed method achieves competitive or superior performance compared with prior offline RL methods, especially in sparse-reward tasks such as Kitchen and AntMaze. We also empirically demonstrate the effectiveness of the anti-noise preference optimization.

Offline reinforcement learning (RL) aims to learn optimal policies from previously collected datasets. Recently, due to their powerful representational capabilities, diffusion models have shown significant potential as policy models for offline RL issues. However, previous offline RL algorithms based on diffusion policies generally adopt weighted regression to improve the policy. This approach optimizes the policy only using the collected actions and is sensitive to Q-values, which limits the potential for further performance enhancement. To this end, we propose a novel preferred-action-optimized diffusion policy for offline RL. In particular, an expressive conditional diffusion model is utilized to represent the diverse distribution of a behavior policy. Meanwhile, based on the diffusion model, preferred actions within the same behavior distribution are automatically generated through the critic function. Moreover, an anti-noise preference optimization is designed to achieve policy improvement by using the preferred actions, which can adapt to noise-preferred actions for stable training. Extensive experiments demonstrate that the proposed method provides competitive or superior performance compared to previous state-of-the-art offline RL methods, particularly in sparse reward tasks such as Kitchen and AntMaze. Additionally, we empirically prove the effectiveness of anti-noise preference optimization.

\end{abstract}

\vspace{-2.0mm}
\section{Introduction}
Offline reinforcement learning (RL) seeks to learn effective policies from previous experiences without interacting with the environment\cite{levine2020offline}. Avoiding costly or risky online interactions makes offline RL appealing for numerous real-world applications. Extending online RL methods to the offline setting primarily faces the issue of distribution shift \cite{fujimoto2019off}. Existing offline RL methods predominantly concentrate on addressing this issue by employing conservative updates for Q-functions \cite{kumar2020conservative, yu2021combo, kostrikov2021offline}, constraining a policy to stay close to the behavior policy via weighted regression (WR) \cite{nair2020awac, fujimoto2021minimalist}, or integrating the two approaches \cite{wang2020critic, ma2024mutual}. However, most offline RL methods usually parameterize a policy as an unimodal Gaussian model with the learned mean and variance. This scheme may not be feasible when the collected data distribution is complex and diverse. Moreover, the collected behaviors can be highly diverse in the real world and present strong multi-modalities \cite{kang2024efficient}. Therefore, more expressive policy models are urgently needed.

Fortunately, diffusion models \cite{ho2020denoising, song2020denoising} are usually used to model diverse or multimodal distributions due to their strong expressiveness, and have achieved the latest benchmark in image generation tasks \cite{dhariwal2021diffusion, rombach2022high}. Recently, the offline RL works using the diffusion model as a policy model have shown great potential \cite{wang2022diffusion} and made good progress \cite{kang2024efficient}. Despite the impressive improvement that these works have achieved, they still adopt the WR method \cite{nair2020awac} for policy improvement. However, the WR method has two critical drawbacks preventing the performance improvement of diffusion policies. \emph{First}, WR aims to help prevent querying out-of-sample actions, which can result in policy performance being greatly limited by the collected data. In most offline tasks, the collected data is only a small part of the whole state-action space. Thus, WR has a performance upper constraint and cannot greatly improve the performance of policies. \emph{Second}, WR typically uses Q-value to calculate the updated weights of policies and is sensitive to Q-value. In offline RL, the estimation of the Q-value generally exits error due to the out-of-distribution (OOD) problem \cite{ma2021conservative}. Consequently, WR sensitive to Q-value has unstable training, which can degrade performance.

Recently, RL with preference-based reward learning has shown great potential for further improving the performance of large language models \cite{ouyang2022training}. Meanwhile, direct preference optimization (DPO) \cite{rafailov2024direct} deduces that a simple classification loss using preference models can be directly used to solve the RL problem for policy improvement. Inspired by this, we propose a novel \textbf{P}referred-\textbf{A}ction-\textbf{O}ptimized \textbf{D}iffusion \textbf{P}olicy (PAO-DP) for offline RL. PAO-DP focuses on utilizing preferred actions to optimize the diffusion policy via a preference model (e.g., Bradley-Terry \cite{bradley1952rank}), instead of WR. To our knowledge, PAO-DP is the first attempt to integrate preference optimization with diffusion models for offline RL. Specifically, PAO-DP mainly has the following characteristics: 1) The diverse distribution of the behavior policy is represented using a conditional diffusion model by the behavior cloning approach. 2) Preferred actions are automatically generated by the critic function based on the diffusion model, rather than manually annotating. All preferred actions are sampled from the same distribution of the behavior policy for accuracy evaluation. 3) An anti-noise preference optimization is designed to improve the diffusion policy further by using preferred actions, while also handling noise-preferred actions for stable training. We test PAO-DP on the popular D4RL benchmarking \cite{fu2020d4rl}. The experiment results demonstrate the proposed method achieves competitive or superior performance compared with prior offline RL methods, especially in sparse-reward tasks such as Kitchen and AntMaze.

%最近的offline rl情况
%引出diffusion
%引出diffusion提高的问题

% 权重回归，权重需要自己设计，参数难调
% 权重回归，只会对样本里的动作进行微调，性能受限
% 权重回归，Q值直接用来计算权重系数，策略性能对Q值较为敏感。

\vspace{-2.0mm}
\section{Preliminaries}

%\subsection{Problem Settings only with Label Signals}
\subsection{Offline Reinforcement Learning}

%\mathcal{S}\times\mathcal{A}\times\mathcal{S} \rightarrow [0,1]

\textbf{Offline RL Settings.} A decision-making problem is usually modeled as a Markov decision process (MDP). The MDP is defined as a tuple~\cite{puterman2014markov}: ($\mathcal{S}$, $\mathcal{A}$, $\mathcal{P}$, $r$, $\gamma$), where $\mathcal{S}$ and $\mathcal{A}$ denote the state and action spaces respectively, $\mathcal{P}(\bm{s}'|\bm{s},\bm{a}): \mathcal{S}\times\mathcal{A}\times\mathcal{S} \rightarrow [0,1]$ denotes the transition probability from state $\bm{s}$ to the next state $\bm{s}'$ after taking the action $\bm{a}$, $r: \mathcal{S}\times\mathcal{A} \rightarrow \mathbb{R}$ represents the reward function, $\gamma \in [0,1)$ is the discount factor.  The goal is to learn a policy $\pi(\bm{a}|\bm{s})$ that maximizes the cumulative discounted reward $\sum_{t=0}^{\infty}\gamma^{t}r(\bm{s}_{t},\bm{a}_{t})$. 
% The Q-function of the policy is usually defined as $Q(\bm{s}_{t},\bm{a}_{t}) = \mathbb{E}_{\bm{a}_{t+1},\bm{a}_{t+2}\sim \pi}[\sum_{c=0}^{\infty}\gamma^{c}r(\bm{s}_{t+c+1},\bm{a}_{t+c+1})]$. 
% In offline settings, a static dataset $\mathcal{D} \triangleq \{(\bm{s},\bm{a},r,\bm{s}^{\prime})\}$, collected through a behavior policy $\pi^{b}$, is provided in place of the environment. Offline RL algorithms aim to learn an effective policy completely from the static dataset, without online interacting with the environment. 
In an offline setting, the objective is to learn a policy that maximizes reward from a static sample dataset $\mathcal{D}$ collected by an unknown behavioral policy $\pi^{b}$, without further online interaction with the environment~\cite{chen2020bail, agarwal2020optimistic}.

% In an offline setting, the objective is to learn a policy that maximizes reward from static sample data collected by an unknown behavioral policy π, without further engaging in online interaction with the environment.

% Offline RL algorithms typically tackle the distribution shift \cite{fujimoto2019off} issue caused by the dataset $\mathcal{D}$ by either modifying the policy evaluation step to conserve Q-learning or directly constraining policy improvement via weighted regression.

\textbf{Weighted Regression.} In offline RL algorithms, due to the importance of adhering to the behavior policy $\pi^{b}$, previous works \cite{peng2019advantage, nair2020awac} explicitly constraint the learned policy $\pi$ to be closed to $\pi^{b}$, while maximizing the expected value of the Q-function \cite{chen2022offline}:
% \small
% \begin{equation}
%     \label{e_1}
%     % J(\pi)=\arg \max_{\pi} \mathbb{E}_{\bm{s},\bm{a}}\left[\pi(\bm{a}|\bm{s}) Q_{\phi}(\bm{s},\bm{a})\right] - \frac{1}{\eta}\mathbb{E}_{\bm{s}} \left[ D_{KL}(\pi(\cdot|\bm{s})||\pi^{b}(\cdot|\bm{s}))\right].\\
%     J(\pi)=\arg \max_{\pi} \mathbb{E}_{\bm{s}}\left[\sum_{\bm{a}}\pi(\bm{a}|\bm{s}) Q_{\phi}(\bm{s},\bm{a})\right] - \frac{1}{\eta}\mathbb{E}_{\bm{s}} \left[ D_{KL}(\pi(\cdot|\bm{s})||\pi^{b}(\cdot|\bm{s}))\right].
% \end{equation}
% \normalsize
\small
\begin{equation}
    \label{e_1}
    J(\pi)=\arg \max_{\pi} \mathbb{E}_{\bm{s}}\left[\sum_{\bm{a}}\pi(\bm{a}|\bm{s}) Q_{\phi}(\bm{s},\bm{a})\right] - {\eta}\mathbb{E}_{\bm{s}} \left[ D_{KL}(\pi(\cdot|\bm{s})||\pi^{b}(\cdot|\bm{s}))\right].
\end{equation}
\normalsize
The first term corresponds to the optimization objective of the policy $\pi$, in which $Q_{\phi}$ is a learned Q-function of $\pi$. The second term acts as a regularization term to constrain the learned policy within the bounds of the dataset $\mathcal{D}$ with $\eta$ being the coefficient. By solving the Eq.~(\ref{e_1}) for the extreme points of the policy $\pi$, the closed-form solution for the optimal policy $\pi^{*}$ can be obtained~\cite{nair2020awac}:
% According to the Equation~\ref{e_1}, the optimal policy $\pi^{*}$ can be derived by utilizing the Lagrange multiplier \cite{nair2006accelerating}:
\small
\begin{equation} 
    \label{e_2}
    \pi^{*} = \frac{1}{Z(\bm{s})}\pi^{b}(\bm{a}|\bm{s})\exp(\frac{1}{\eta} Q_{\phi}(\bm{s},\bm{a})),\\
\end{equation}
\normalsize
where $Z(\bm{s})$ is the partition function. Eq.~(\ref{e_2}) can be regarded as a policy improvement step. Due to the difficulty in modeling the behavior policy $\pi^{b}$ and the problem of inaccurate estimation of the Q-function, it is challenging to sample and obtain the optimal action directly. The existing methods \cite{peng2019advantage, wang2020critic, chen2020bail, chen2022offline} 
typically project the optimal policy $\pi^{*}$ onto a parameterized policy $\pi_{\theta}$ to bypass this issue:
\small
\begin{equation}\label{e2-1}
\begin{split} 
 \mathop{\arg\min}\limits_{\bm{\theta}} \; \mathbb{E}_{\bm{s} \sim \mathcal{D}}[D_{KL}(\pi^{*}(\cdot|\bm{s})||\pi_{\theta}(\cdot|\bm{s}))]  = \mathop{\arg\max}\limits_{\bm{\theta}} \; \mathbb{E}_{(\bm{s}, \bm{a}) \sim \mathcal{D}}\left[\frac{1}{Z(\bm{s})}\log \pi_{\theta}(\bm{a}|\bm{s})\exp(\frac{1}{\eta} Q_{\phi}(\bm{s},\bm{a}))\right].
\end{split}
\end{equation}
\normalsize
% \begin{equation}\label{e2-1}
% \begin{split} 
% & \mathop{\arg\min}\limits_{\bm{\theta}} \; \mathbb{E}_{\bm{s} \sim \mathcal{D}}[D_{KL}(\pi^{*}(\cdot|\bm{s})||\pi_{\theta}(\cdot|\bm{s}))] = \mathop{\arg\max}\limits_{\bm{\theta}} \; \mathbb{E}_{s\sim \mathcal{D}}\mathbb{E}_{\pi^{*}(\cdot|s)}\log\pi_{\theta}(\cdot|s).
% \end{split}
% \end{equation}
% \normalsize
Such a method is commonly known as weighted regression, where $\exp(\frac{1}{\eta} Q_{\phi}(\bm{s},\bm{a}))$ serves as the regression weights.

\subsection{Diffusion Probabilistic Model}

Diffusion-based generative models \cite{song2019generative, ho2020denoising} mainly include forward and reverse diffusion processes. In particular, suppose there is a real data distribution $q(x)$, from which a sample $x^{0} \sim q(x)$ is drawn. The forward diffusion process followed a Markov chain gradually adds Gaussian noise to the sample through $K$ incremental steps, yielding a sequence of noisy samples $x^{1}, ..., x^{K}$ controlled by a pre-defined variance schedule $\beta^{1},...,\beta^{K}$. The forward process is expressed as: 
\small
\begin{equation} 
    q(\bm{x}^{1:K}|\bm{x}^{0}) = \prod_{k=1}^{K}q(\bm{x}^{k}|\bm{x}^{k-1}), \quad q(\bm{x}^{k}|\bm{x}^{k-1}) = \mathcal{N}(\bm{x}^{k};\sqrt{1-\beta^{k}}\bm{x}^{k-1},\beta^{k}\bm{I}).
\end{equation}
\normalsize
In the reverse diffusion processes, diffusion models mainly learn a conditional distribution $p_{\theta}(\bm{x}^{k-1}|x^{k})$ for generating new samples. The reverse process is represented as: 
\small
\begin{equation} 
p_{\theta}(\bm{x}^{0:K}) = p(\bm{x}^{K})\prod_{k=1}^{K}p_{\theta}(\bm{x}^{k-1}|\bm{x}^{k}), \quad p_{\theta}(\bm{x}^{k-1}|\bm{x}^{k})=\mathcal{N}(\bm{x}^{k-1};\bm{\mu}_{\theta}(\bm{x}^{k},k),\bm{\Sigma}_{\theta}(\bm{x}^{k},k)),
\end{equation}
\normalsize
where $p(\bm{x}^{K}) = \mathcal{N}(\bm{0},\bm{I})$. The diffusion training is executed by maximizing the evidence lower bound (ELBO): 
% $\mathbb{E}_{q}[\ln\frac{p_{\theta}(\bm{x}^{0:K})}{q(\bm{x}^{1:K}|\bm{x}^{0})}]$
$\mathbb{E}_{q}[\ln {p_{\theta}(\bm{x}^{0:K})}-\ln q(\bm{x}^{1:K}|\bm{x}^{0})]$ \cite{blei2017variational,kang2024efficient}.

\vspace{-2.0mm}
\section{Method}
%we propose a novel method to further improve policy performance based on diffusion policy.

In this section, we detail the overall design of the proposed preferred-action optimized diffusion policy (PAO-DP).
Firstly, we formulate an RL behavior policy using a conditional diffusion model. Then, the preferred-action automatic generation module is presented. The module aims to utilize the behavior policy and critic function to generate preferred actions. Finally, the policy anti-noise preference optimization is given to achieve policy improvement by using the preferred actions.

%as shown in Figure. \ref{fig:overall_method}.

\subsection{Conditional Diffusion Policy}
Generally, the reverse process of a conditional diffusion model is used as an RL parametric policy \cite{wang2022diffusion, kang2024efficient}:
\small
\begin{equation} 
    \label{e_cdp_1}
    \pi^{b}_{\theta}(\bm{a}_{t}|\bm{s}_{t}) = p_{\theta}(\bm{a}^{0:K}_{t}|\bm{s}_{t}) = p(\bm{a}^{k}_{t})\prod_{k=1}^{K}p_{\theta}(\bm{a}_{t}^{k-1}|\bm{a}_{t}^{k},\bm{s}_{t}),\\
\end{equation}
\normalsize
where $k \in \{1,...,K\}$ denotes the diffusion timestep, $t\in \{1,...,T\}$ represents the trajectory timestep, $\bm{a}^{K} \sim \mathcal{N}(\mathbf{0},\mathbf{I})$. Herein, the end sample $\bm{a}^{0}$ of the reverse chain is the action for the RL policy. $p_{\theta}(\bm{a}_{t}^{k-1}|\bm{a}_{t}^{k},\bm{s}_{t})$ can usually be modeled as a Gaussian distribution. Following the denoising diffusion probabilistic model (DDPM) \cite{ho2020denoising},  $p_{\theta}(\bm{a}_{t}^{k-1}|\bm{a}_{t}^{k},\bm{s}_{t})$ is parameterized as a noise prediction model with a fixed covariance matrix: $\mathbf{\Sigma}_{\theta}(\bm{a}^{k}_{t},k;\bm{s}_{t}) = \eta^{k}\mathbf{I}$ and mean built as:
\small
\begin{equation} 
    \label{e_cdp_2}
    \mathbf{\mu}_{\theta}(\bm{a}_{t}^{k},k;\bm{s}_{t}) = \frac{1}{\sqrt{\alpha_{k}}}\left(\bm{a}^{k}_{t}-\frac{\beta_{k}}{\sqrt{1-\bar{\alpha}_{k}}}\bm{\epsilon}_{\theta}(\bm{a}^{k}_{t},k; \bm{s}_{t})\right),\\
\end{equation}
\normalsize
where $\alpha^{k} = 1-\beta^{k}$, $\bar{\alpha}^{k} = \prod_{i=1}^{k}\alpha^{i}$, $\bm{\epsilon}_{\theta}$ is a parametric noise model. In DDPM, we can obtain an action through the following sampling process:
\small
\begin{equation} 
    \label{e_cdp_3}
    \bm{a}^{k-1}_{t} = \frac{\bm{a}^{k}_{t}}{\sqrt{\alpha^{k}}} - \frac{\beta^{k}}{\sqrt{\alpha(1-\bar{\alpha}^{k})}}\bm{\epsilon}(\bm{a}^{k}_{t},k;\bm{s}_{t}) + \sqrt{\beta_{i}}\bm{\epsilon}, \quad \bm{\epsilon} \sim \mathcal{N}(0,I), \quad for \,\:  k= K,...,1.
\end{equation}
\normalsize
Similar to DDPM,  the objective proposed by \cite{ho2020denoising} can be simplified to train the $\bm{\epsilon}$ model by:
\small
\begin{equation} 
    \label{e_cdp_4}
    \mathcal{L}_{d}(\theta) = \mathbb{E}_{k \sim \mathcal{U},\bm{\epsilon} \sim \mathcal{N}(0,I),(\bm{s}_{t},\bm{a}_{t}^{0})\sim \mathcal{D}}\left[||\bm{\epsilon}-\bm{\epsilon}_{\theta}\left(\sqrt{\bar{\alpha}_{k}}\bm{a}_{t}^{0} + \sqrt{1-\bar{\alpha}_{k}}\bm{\epsilon},k;\bm{s}_{t}\right) ||^{2}\right],
\end{equation}
\normalsize
where $\mathcal{U}$ is a uniform distribution over $\{1,...,K\}$ and $\mathcal{D}$ denotes the offline dataset collected by a behavior policy $\pi^{b}$. The diffusion loss $\mathcal{L}_{d}(\theta)$ aims to learn the behavior policy through the conditional diffusion model with $\bm{\epsilon}_{\theta}$, and it is just a loss of behavior cloning. Herein, $\pi^{b}_{\bm{\epsilon}_{\theta}}(\bm{a}_{t}|\bm{s}_{t})$ is used to represent the learned diffusion behavior policy.

\subsection{Preferred-Action Automatic Generation}

Based on the learned behavior policy, we further use preference optimization to achieve policy improvement, instead of weighted regression. One of the keys to preference optimization is to obtain preferred actions. In the field of preference learning \cite{hejna2024inverse}, preference data is generally provided in advance. However, this requires a lot of manpower and material resources. To avoid this issue, we design an automatic acquisition approach that relies on the Q-function. Specifically, we need to automatically generate preferred-action samples, i.e., $\{\bm{s}_{t},\bm{a}_{t},\hat{\bm{a}}_{t},\Gamma_{Q(\bm{s}_{t},\bm{a}_{t})>Q(\bm{s}_{t},\hat{\bm{a}}_{t})}\} \in \mathcal{D}_{p}$, where $\bm{a}_{t}$ denotes the behavior action corresponding to the state $\bm{s}_{t}$ in the offline data $\mathcal{D}$, $\hat{\bm{a}}_{t}$ is another action obtained through the following sampling method, 
% $\Gamma_{Q(\bm{s}_{t},\bm{a}_{t})>Q(\bm{s}_{t},\bm{a}_{t})}$ is a symbolic function used to determine which action has a larger Q-value:
$\Gamma_{Q(\bm{s}_{t},\bm{a}_{t})>Q(\bm{s}_{t},\hat{\bm{a}}_{t})}$ is a symbolic function used to determine which action has a larger Q-value:
% \begin{small}
% \begin{align} 
% \label{e_paag_1}
% \begin{split}
% \Gamma_{Q(\bm{s}_{t},\bm{a}_{t})>Q(\bm{s}_{t},\bm{a}_{t})}= \left \{
% \begin{array}{rr}
%     1,                    & Q(\bm{s}_{t},\bm{a}_{t})>Q(\bm{s}_{t},\bm{a}_{t})\\
%     -1,                                 & otherwise
% \end{array}
% \right.
% \end{split}.
% \end{align}
% \end{small}
\begin{small}
\begin{align} 
\label{e_paag_1}
    \begin{split}    \Gamma_{Q(\bm{s}_{t},\bm{a}_{t})>Q(\bm{s}_{t},\hat{\bm{a}}_{t})}= \left \{
    \begin{array}{rr}
        1,                    & Q(\bm{s}_{t},\bm{a}_{t})>Q(\bm{s}_{t},\hat{\bm{a}}_{t})\\
        -1,                                 & otherwise
    \end{array}
    \right.
    \end{split}.
\end{align}
\end{small}
Meanwhile, the Q-function $Q_{\phi}(\bm{s}_{t},\bm{a}_{t})$ parameterized by $\phi$ can be learned by utilizing the existing Q-learning framework, e.g., TD3 \cite{fujimoto2018addressing} and implicit Q-learning (IQL) \cite{kostrikov2021offline}

\textbf{Preferred-Action Sampling.} Considering the OOD problem in Q-value estimation, we hope another action $\hat{\bm{a}}_{t}$ can be obtained from the distribution of the behavior policy. Due to the learned policy $\pi^{b}_{\bm{\epsilon}_{\theta}}$ modeling the distribution of the behavior policy, $\hat{\bm{a}}_{t}$ can be sampled from $\pi^{b}_{\bm{\epsilon}_{\theta}}(\hat{\bm{a}}_{t}|\bm{s}_{t})$ directly. Generally, higher-quality actions are beneficial for the improvement of policies, especially in preference learning. This further motivates us to sample the action $\hat{\bm{a}}_{t}$ from the optimal policy $\pi^{*}$.

Since the exact and derivable density function of $\pi^{*}$ cannot be determined, it is difficult to directly draw samples from $\pi^{*}$. Drawing on insights from \cite{chen2022offline} and according to Eq.~(\ref{e_2}), the optimal policy $\pi^{*}$ can be further expressed as:
\small
\begin{equation} 
    \label{e_paag_2}
    \pi^{*}(\bm{a}_{t}|\bm{s}_{t})\propto \pi^{b}_{\bm{\epsilon}_{\theta}}(\bm{a}_{t}|\bm{s}_{t})\exp(\eta Q_{\phi}(\bm{s}_{t},\bm{a}_{t})),\\
\end{equation}
\normalsize
Based on this formula, we use an importance sampling technique to sample actions. In particular, for each state $\bm{s}_{t}$, we firstly draw $N$ actions from the learned behavior policy $\pi^{b}_{\bm{\epsilon}_{\theta}}$. Then, these actions are evaluated with the learned Q-function $Q_{\phi}$. Finally, an action is drawn from $N$ actions with $\exp(\eta Q_{\phi}(\hat{\bm{a}}_{t},\bm{s}_{t})$ being the sampling weights \cite{chen2022offline}. Additionally, we can use greedy sampling to obtain $\hat{\bm{a}}_{t}$ from $N$ actions.

% To automatically generate preferred-action samples about the expected cumulative return, i.e., Q value. Specifically, in the same state $\bm{s}_{t}$, we would like to obtain two actions and their Q values to select the preferred action with a higher Q. Herein, one of the two actions is a sample action $\bm{a}_{t}$. 

\subsection{Policy Anti-Noise Preference Optimization}
\label{Policy Anti-Noise Preference Optimization}

% instead of just adjusting the probability of sample actions from the fixed dataset $\mathcal{D}$ like weighted regression. The preferred actions are not limited to the fixed dataset but are within the sample distribution, which promotes the policy with more possibilities for improvement.

After obtaining the preferred action samples, the behavior policy $\pi^{b}_{\bm{\epsilon}_{\theta}}$ can be further improved stably by anti-noise preference optimization. Specifically, firstly, according to Eq.~(\ref{e_2}), we can obtain an expression for Q-function:
\small
\begin{equation} 
    \label{e_po_1}
    Q_{\phi}(\bm{s}_{t},\bm{a}_{t})) = \eta \log \frac{\pi^{*}(\bm{a}_{t}|\bm{s}_{t})}{\pi^{b}_{\bm{\epsilon}_{\theta}}(\bm{a}_{t}|\bm{s}_{t})}+\eta \log Z(s).\\
\end{equation}
\normalsize

Then, based on $\{\bm{s}_{t},\bm{a}_{t},\hat{\bm{a}}_{t},\Gamma_{Q(\bm{s}_{t},\bm{a}_{t})>Q(\bm{s}_{t},\hat{\bm{a}}_{t})}\}$, the popular Bradley-Terry (BT) \cite{bradley1952rank, rafailov2024direct} model is used to model the action preference distribution $p^{*}$:
\small
\begin{equation} 
    \label{e_po_2}
    p^{*}(\hat{\bm{a}}_{t} \succ \bm{a}_{t} | \bm{s}_{t}) = \frac{\exp (Q_{\phi}(\bm{s}_{t},\hat{\bm{a}}_{t}))}{\exp (Q_{\phi}(\bm{s}_{t},\hat{\bm{a}}_{t})) + \exp (Q_{\phi}(\bm{s}_{t},\bm{a}_{t}))}.\\
\end{equation}
\normalsize

Putting the expression of Q in Eq.~(\ref{e_po_1}) into Eq.~(\ref{e_po_2}), we can obtain the action preference probability in terms of only the optimal policy $\pi^{*}$ and the behavior policy $\pi^{b}_{\bm{\epsilon}_{\theta}}$, expressed as \cite{rafailov2024direct}:
{
%\small
\begin{equation}
    \label{e_po_3}
    p^{*}(\hat{\bm{a}}_{t} \succ \bm{a}_{t} | \bm{s}_{t}) = \frac{1}{1+\exp \left(\eta \log \frac{\pi^{*}(\bm{a}_{t}|\bm{s}_{t})}{\pi^{b}_{\bm{\epsilon}_{\theta}}(\bm{a}_{t}|\bm{s}_{t})}-\eta \log \frac{\pi^{*}(\hat{\bm{a}}_{t}|\bm{s}_{t})}{\pi^{b}_{\bm{\epsilon}_{\theta}}(\hat{\bm{a}}_{t}|\bm{s}_{t})}\right)}.\\
\end{equation}
% \normalsize
}
% The derivation is in the Appendix \ref{Deriving the Preference Optimization Objective of PAO-DP}. Herein, we parameterize the optimal policy by $\psi$, represented as the surrogate optimal policy $\pi_{\psi}$, where the structure of $\pi_{\psi}$ is the same as $\pi^{b}_{\bm{\epsilon}_{\phi}}$, and $\pi_{\psi}$ is further optimized based on $\pi^{b}_{\bm{\epsilon}_{\phi}}$ for the policy improvement. Subsequently, we can formulate a maximum likelihood objective for $\pi_{\psi}$. The policy improvement objective can be expressed as:
The derivation is in the Appendix \ref{Deriving the Preference Optimization Objective of PAO-DP}. Herein, we parameterize the optimal policy by $\psi$, represented as the surrogate optimal policy $\pi_{\psi}$, where the structure of $\pi_{\psi}$ is the same as $\pi^{b}_{\bm{\epsilon}_{\theta}}$, and $\pi_{\psi}$ is further optimized based on $\pi^{b}_{\bm{\epsilon}_{\theta}}$ for the policy improvement. Subsequently, we can formulate a maximum likelihood objective for $\pi_{\psi}$. The policy improvement objective can be expressed as:
\begin{equation}
    \label{e_po_4}
    \!\!\mathcal{L}_{imp}(\psi;\theta)\!\! =\!\! -\mathbb{E}_{(\bm{s}_{t},\bm{a}_{t},\hat{\bm{a}}_{t},\Gamma) \in \mathcal{D}_{p}} \left[ \log \sigma \left((\eta * \Gamma) \log \frac{\pi_{\psi}(\hat{\bm{a}}_{t}|\bm{s}_{t})}{\pi^{b}_{\bm{\epsilon}_{\theta}}(\hat{\bm{a}}_{t}|\bm{s}_{t})}-(\eta * \Gamma) \log \frac{\pi_{\psi}(\bm{a}_{t}|\bm{s}_{t})}{\pi^{b}_{\bm{\epsilon}_{\theta}}(\bm{a}_{t}|\bm{s}_{t})}\right)\right],\\
\end{equation}
where $\sigma$ is the logistic function. Furthermore, considering the imprecision possibility of estimating Q-values in offline RL, such inaccuracies may introduce noises in the preference action samples. This may lead to instability in training during the policy improvement process. To alleviate this issue, we further establish an anti-noise optimization objective. Specifically, assume that the preferred action label may be flipped with a small probability $\lambda \in (0,0.5)$ and $\hat{\bm{a}}_{t}$ is preferred $\bm{a}_{t}$, we can obtain a conservative preferred-action probability $p^{*}(\hat{\bm{a}}_{t} \succ \bm{a}_{t} | \bm{s}_{t} ) = 1- \lambda$ \cite{cdpo}. Hence, we can calculate a cross-entropy loss as the anti-noise optimization objective \cite{chowdhury2024provably}:
\begin{equation}
    \label{e_po_5}
    \mathcal{L}_{anti}(\psi;\theta) = (1-\lambda) \mathcal{L}_{imp}(\psi;\theta) + \lambda \mathcal{L}_{imp}^{-\Gamma}(\psi;\theta),
\end{equation}
where $\mathcal{L}_{imp}^{-\Gamma}(\psi;\theta)$ represents the objective whose preference actions have been flipped.

In our paper, to train more efficiently, we adopt a strategy of simultaneously training behavior policy $\pi^{b}_{\bm{\epsilon}_{\theta}}$ and surrogate optimal policy $\pi_{\psi}$, rather than dividing the training into two stages. Hence, the loss function of the surrogate optimal policy is a linear combination of policy cloning and policy improvement, represented as $\mathcal{L}(\psi) = \mathcal{L}_{d}(\psi) + \xi \mathcal{L}_{anti}(\psi;\theta) $, where $\xi$ is a hyperparameter to provide a weight for the policy improvement part. This parameter can also be used to measure the effect of the policy improvement.  The overall algorithm is given in Algorithm~\ref{alg:our_algorithm}
\begin{small}
\begin{algorithm}[htbp] 
    \caption{PAO-DP Algorithm}
    \label{alg:our_algorithm}    
    \KwIn{ Dataset $ \mathcal{D} =\{(\bm{s}_t,\bm{a}_t,r_t,\bm{s}_{t+1})_{i=0}^{n}\}$ }
    \KwOut{Surrogate optimal policy $\pi_{\psi}$, Behavior policy $\pi^{b}_{\bm{\epsilon}_{\theta}}$}
    \For{each batch}{
        Sample a batch of transitions $(\bm{s}_t,\bm{a}_t,r_t,\bm{s}_{t+1})$ from the buffer $\mathcal{D}$\\
        \textcolor{blue}{\# Behavior policy learning}\\
        Update $\bm{\epsilon}_{\theta}$ via the Eq.~(\ref{e_cdp_4})\\
        \textcolor{blue}{\# Critic learning}\\
        Update $Q_{\phi}$ via the existing Q-learning method, e.g., IQL\\
        \textcolor{blue}{\# Preferred-action automatic generation}\\
        Sample $N$ actions in $\bm{s}_{t}$ by the behavior policy $\pi^{b}_{\bm{\epsilon}_{\theta}}$ using the Eq.~(\ref{e_cdp_3})\\
        Select an action from $N$ actions via the importance sampling using  Eq.~(\ref{e_paag_2})\\
        Generate preferred-action sample $\{\bm{s}_{t},\bm{a}_{t},\hat{\bm{a}}_{t},\Gamma_{Q(\bm{s}_{t},\bm{a}_{t})>Q(\bm{s}_{t},\hat{\bm{a}}_{t})}\}$ via Eq.~(\ref{e_paag_1})\\
        \textcolor{blue}{\# Policy improvement}\\
        Calculate the anti-noise preference loss $\mathcal{L}_{anti}(\psi;\theta)$ via  Eq.~(\ref{e_po_5})\\ 
        Calculate the diffusion loss $\mathcal{L}_{d}(\psi)$ via Eq.~(\ref{e_cdp_4})\\
        Update the surrogate optimal policy $\pi_{\psi}$ via the total loss function: $\mathcal{L}(\psi) = \mathcal{L}_{d}(\psi) + \xi \mathcal{L}_{anti}(\psi;\theta)$\\
        % $\pi_{\thet\bm{a}_{t}}\leftarrow \tau\pi_{\theta}+(1-\tau)\pi_{\thet\bm{a}_{t}}$\;
    }
\end{algorithm}
\vspace{-0.2cm}
\end{small}

\vspace{-2.0mm}
\section{Experiments}
We conduct extensive experiments on the popular D4RL benchmark \cite{fu2020d4rl} to validate the effectiveness of the proposed method, especially in sparse-reward tasks. 
% Further, ablation studies are performed to analyze the contribution of the main components of PAO-DP. Besides, we also conduct parameter sensitivity analysis on the different methods.
Additionally, we performed ablation studies to dissect the contributions of the main components of PAO-DP. Furthermore, we carried out a parameter sensitivity analysis across different methods to assess their robustness and performance.

\subsection{Experimental Settings}
% \textbf{Datasets.} The proposed method is evaluated in four different domains in D4RL, including Kitchen, AntMaze, Adroit, and Gym-locomation. In our experiments, Kitchen and AntMaze are all composed of sparse-reward tasks, while Adroit and Gym-locomation are not. In the kitchen domain, the agent must accomplish four sequential subtasks to reach a goal state configuration. This is a highly challenging long-horizon task scenario. In the AntMaze domain, the agent is required to stitch various suboptimal trajectories to find the path to the maze goal. Adroit consists of high-dimensional robotic manipulation tasks, where the state-action region reflected via the offline data is usually very narrow. Gym-MuJoCo locomotion contains many standard tasks, in which the reward function is quite smooth.
\textbf{Baselines.} The proposed method is evaluated in four different domains in D4RL, including Kitchen, AntMaze, Adroit, and Gym-locomation. In our experiments, Kitchen and AntMaze are all composed of sparse-reward tasks, while Adroit and Gym-locomation are not. For each domain, we consider different classes of baselines for providing comprehensive evaluation. For policy regularization-based methods, the simple method is the canonical behavior cloning (BC) baseline. AWAC \cite{nair2020awac} and CRR \cite{wang2020critic} achieve policy improvement by adding advantage-based weights in the policy loss function. For Q-value constraint methods, CQL \cite{kumar2020conservative} and IQL \cite{kostrikov2021offline} adopt conservative Q-function updates or expectile regression for constraining the policy evaluation process. For sequence modeling approaches, we consider the Decision Transformer (DT) \cite{chen2021decision} baseline. For diffusion policies, the diffusion Q-learning (DIFF-QL) \cite{wang2022diffusion} baseline first represents the policy as a diffusion model, and couples the behavior cloning and policy improvement using WR. Based on DIFF-QL, the efficient diffusion policy (EDP) \cite{kang2024efficient} method approximately constructs actions from corrupted ones for realizing efficient training. CRR+EDP and IQL+EDP respectively adopt CRR and IQL based on EDP for policy improvement. CRR+EDP focuses on advantage-based weight regression, while IQL+EDP uses Q-based weight regression. The SfBC \cite{chen2022offline} baseline decouples the learned policy into a diffusion behavior model and an action evaluation model, without a policy improvement process.

\textbf{Experimental Details.} 
% In our experiments, following EDP, the critic function network architecture, which is a three-layer MLP with Mish \cite{misra2019mish} activation function, is the same for all the tasks. 
In our experiments, consistent with the EDP framework, we utilize a critic function network architecture comprising a three-layer MLP, employing the Mish activation function \cite{misra2019mish}. This architecture is uniformly applied across all tasks.
To obtain a more accurate estimate of the Q-value from offline data, the critic function is updated through the IQL approach using expectile regression. The noise prediction network $\bm{\epsilon}_{\theta}(\bm{a}^{k}_{t},k;\bm{s}_{t})$ of the diffusion policy initially employs sinusoidal embedding \cite{vaswani2017attention} to encode timestep $k$, subsequently concatenating it with the noisy action $\bm{a}^{k}_{t}$ and the conditional state $\bm{s}_{t}$. Similar to EDP, the policy is trained for 2000 epochs on the Gym-locomation domain and 1000 epochs on the other domains. Every epoch is made up of 1000 policy updates with a batch size of 256. The third-order version of DPM-Solver \cite{lu2022dpm} is used to perform the denoising process of the diffusion policy with 15 steps. For preferred-action automatic generation, we found that greedy strategy sampling performed outstandingly in most tasks, so we used it. The number of sampled actions is 10, i.e., $N=10$. Throughout our paper, the results are reported by averaging 3 random seeds. In addition, all the hyperparameters are listed in the Appendix \ref{Hyperparameters}.

\textbf{Evaluation Metrics.} We evaluate the proposed PAO-DP method using two metrics: Online Model Selection (OMS) and Running Average at Training (RAT) \cite{kang2024efficient}. OMS, introduced by Diffusion-QL \cite{wang2022diffusion}, selects the best-performing model during training, indicating the algorithm's potential. 
In contrast, RAT \cite{kang2024efficient} calculates the running average of evaluation performance over ten consecutive checkpoints and reports the final score, providing a more stable and realistic assessment of an algorithm's performance throughout training. Besides, our paper uses the average normalized score of D4RL to evaluate different methods.

\vspace{-0.2cm}
\subsection{Evaluation Results}
We compare our proposed PAO-DP method with the baseline methods across four task domains, with the results detailed in Table \ref{table:compare_1_four}. The domain-specific analyses are as follows:

\textbf{Results for Kitchen Domain:} 
The Kitchen domain presents a challenging long-term task requiring the agent to accomplish a series of sequential subtasks to reach a goal state configuration. This domain is characterized by sparse rewards and the complexity of achieving the final goal through intermediate steps, making long-term value optimization crucial.
The proposed PAO-DP method significantly outperforms the baselines in the Kitchen domain across all three datasets: \textit{Complete}, \textit{Partial}, and \textit{Mixed}. In the \textit{Complete} task, our method achieves an impressive score of 87.2, which is notably higher than the next-best performance of 75.5 by IQL+EDP using WR. The average performance in the Kitchen domain shows PAO-DP at 67.5, demonstrating a substantial improvement over all other methods. These results underscore the effectiveness of PAO-DP in optimizing long-term value in environments with sparse rewards and complex sequential tasks. By integrating policy improvement using preference optimization with diffusion policies, PAO-DP significantly enhances the agent's performance in this challenging domain, outperforming diffusion policies that incorporate weighted regression.

\textbf{Results for AntMaze Domain:} 
PAO-DP demonstrates superior performance in the AntMaze domain, which is characterized by sparse rewards and challenging sub-optimal trajectory tasks. The average performance across the AntMaze domain for PAO-DP is 79.7, significantly higher than all other baselines. Although CQL marginally surpasses PAO-DP in the Diverse task for AntMaze-\textit{umaze}, scoring 84.0 compared to PAO-DP's 83.7, PAO-DP consistently excels in other settings.
The robustness of PAO-DP is particularly evident in the AntMaze-\textit{medium} and AntMaze-\textit{large} tasks, where other methods, such as BC, AWAC, and DT methods, struggle to perform effectively.

\begin{table}[t]
\small
\centering
\caption{Average normalized score of the baselines and PAO-OP evaluated using the RAT metric.}
\setlength\tabcolsep{1.5pt}
%\resizebox{1.0\textwidth}{!}{
\begin{tabular}{@{}cccccccccc@{}}
\toprule
\textbf{Dataset}               & \textbf{Environment}           & \textbf{BC}    & \textbf{AWAC} & \textbf{CQL}   & \textbf{IQL}   & \textbf{DT}    & \textbf{CRR+EDP} & \textbf{IQL+EDP} & \textbf{PAO-DP (Ours)} \\ \midrule
Complete              & Kitchen               & 33.8  & 39.3 & 43.8  & 62.5  &--       & 73.9    & 75.5    & $\bm{87.2}$ $\pm$ 1.05  \\
Partial               & Kitchen               & 33.8  & 36.6 & 49.8  & 46.3  &--       & 40.0    & 46.3    & $\bm{53.3}$ $\pm$ 0.48  \\
Mixed                 & Kitchen               & 47.5  & 22.0 & 51.0  & 51.0  & --      & 46.1    & 56.5    & $\bm{62.2}$ $\pm$ 2.83  \\ 
\midrule
\multicolumn{2}{c}{\textbf{Average   (FrankaKitchen)}} & 38.4  & 32.6 & 48.2  & 53.3  &--       & 53.3    & 59.4    & $\bm{67.5}$ $\pm$ 14.45 \\ \bottomrule
\toprule
\textbf{Dataset}               & \textbf{Environment}           & \textbf{BC}    & \textbf{AWAC} & \textbf{CQL}   & \textbf{IQL}   & \textbf{DT}    & \textbf{CRR+EDP} & \textbf{IQL+EDP} & \textbf{PAO-DP (Ours)} \\ \midrule
Default               & AntMaze-umaze         & 54.6  & 56.7 & 74.0  & 87.5  & 59.2  & 95.9    & 94.2    & $\bm{97.3}$ $\pm$ 1.69  \\
Diverse               & AntMaze-umaze         & 45.6  & 49.3 & $\bm{84.0}$  & 62.2  & 53.0  & 15.9    & 79.0    & 83.7 $\pm$ 1.69  \\
Play                  & AntMaze-medium        & 0.0   & 0.0  & 61.2  & 71.2  & 0.0   & 33.5    & 81.8    & $\bm{85.3}$ $\pm$ 0.47  \\
Diverse               & AntMaze-medium        & 0.0   & 0.7  & 53.7  & 70.0  & 0.0   & 32.7    & 82.3    & $\bm{83.0}$ $\pm$ 2.16  \\
Play                  & AntMaze-large         & 0.0   & 0.0  & 15.8  & 39.6  & 0.0   & 26.0    & 42.3    & $\bm{58.7}$ $\pm$ 1.25  \\
Diverse               & AntMaze-large         & 0.0   & 1.0  & 14.9  & 47.5  & 0.0   & 58.5    & 60.6    & $\bm{68.7}$ $\pm$ 0.94  \\ 
\midrule
\multicolumn{2}{c}{\textbf{Average (AntMaze)}}         & 16.7  & 18.0 & 50.6  & 63.0  & 18.7  & 43.8    & 73.4    & $\bm{79.7}$ $\pm$ 12.10 \\ \bottomrule
\toprule
\textbf{Dataset}               & \textbf{Environment}           & \textbf{BC}    & \textbf{AWAC} & \textbf{CQL}   & \textbf{IQL}   & \textbf{DT}    & \textbf{CRR+EDP} & \textbf{IQL+EDP} & \textbf{PAO-DP (Ours)} \\ \midrule
human                 & pen                   & 25.8  & 15.6 & 37.5  & 71.5  & --      & 70.2    & 72.7    & $\bm{77.6}$ $\pm$ 0.92  \\
cloned                & pen                   & 38.3  & 24.7 & 39.2  & 37.3  & --      & 54.2    & 70.0    & $\bm{83.6}$ $\pm$ 1.75  \\ 
\midrule
\multicolumn{2}{c}{\textbf{Average (Adroit)}}          & 32.1  & 20.2 & 38.4  & 54.4  & --      & 62.1    & 71.3    & $\bm{80.6}$ $\pm$ 3.31  \\ \bottomrule
\toprule
\textbf{Dataset}               & \textbf{Environment}           & \textbf{BC}    & \textbf{AWAC} & \textbf{CQL}   & \textbf{IQL}   & \textbf{DT}    & \textbf{CRR+EDP} & \textbf{IQL+EDP} & \textbf{PAO-DP (Ours)} \\ \midrule
Med-Expert         & HalfCheetah           & 55.2  & 42.8 & $\bm{91.6}$  & 86.7  & 86.8  & 85.6    & 86.7    & 91.0 $\pm$ 0.48  \\
Med-Expert         & Hopper                & 52.5  & 55.8 & 105.4 & 91.5  & $\bm{107.6}$ & 92.9    & 99.6    & 101.1 $\pm$ 1.01 \\
Med-Expert         & Walker                & 107.5 & 74.5 & 108.8 & 109.6 & 108.1 & $\bm{110.1}$   & 109.0   & 109.3 $\pm$ 0.05 \\
Medium                & HalfCheetah           & 42.6  & 43.5 & 44.0  & 47.4  & 42.6  & $\bm{49.2}$    & 48.1    & 47.0 $\pm$ 0.28  \\
Medium                & Hopper                & 52.9  & 57.0 & 58.5  & 66.3  & 67.6  & $\bm{78.7}$    & 63.1    & 57.4 $\pm$ 0.96  \\
Medium                & Walker                & 75.3  & 72.4 & 72.5  & 78.3  & 74.0  & 82.5    & $\bm{85.4}$    & 82.3 $\pm$ 1.33  \\
Med-Replay         & HalfCheetah           & 36.6  & 40.5 & $\bm{45.5}$  & 44.2  & 36.6  & 43.5    & 43.8    & 43.8 $\pm$ 0.04  \\
Med-Replay         & Hopper                & 18.1  & 37.2 & 95.0  & 94.7  & 82.7  & 99.0    & $\bm{99.1}$    & 90.6 $\pm$ 0.69  \\
Med-Replay         & Walker                & 26.0  & 27.0 & 77.2  & 73.9  & 66.6  & 63.3    & $\bm{84.0}$    & 83.9 $\pm$ 0.39  \\ 
\midrule
\multicolumn{2}{c}{\textbf{Average (Locomotion)}}      & 51.9  & 50.1 & 77.6  & 77.0  & 74.7  & 78.3    & $\bm{79.9}$    & 78.5 $\pm$ 22.23 \\ \bottomrule
\end{tabular}
\label{table:compare_1_four}
\vspace{-0.7cm}
\end{table}
\normalsize

\textbf{Results for Adroit Domain:} 
The Adroit domain presents unique challenges due to its datasets being primarily collected through human behavior, resulting in a narrow state-action region in the offline data. In the Adroit domain, PAO-DP once again demonstrates its effectiveness. For the \textit{human} dataset in the Adroit-\textit{pen} environment, PAO-DP achieves a score of 77.6, surpassing IQL+EDP's 72.7 and CRR+EDP's 70.2. The average performance across the Adroit domain for PAO-DP is 80.6, highlighting its superior capability compared to other methods. 
These results demonstrate that PAO-DP can maintain high performance within the narrow operational bounds defined by the offline data. The ability of PAO-DP to maintain expected behaviors while improving performance through preference optimization underscores its advantage in constrained environments over the use of weighted regression for policy improvement.

% The ability to maintain expected behaviors while optimizing performance using preferred actions highlights the strength of PAO-DP in constrained environments, where robust policy regularization is essential.

% These results demonstrate that PAO-DP can maintain high performance within the narrow operational bounds defined by the offline data, making it particularly well-suited for tasks in the Adroit domain.

\textbf{Results for Locomotion Domain:} 
The Locomotion domain contains many standard tasks with relatively smooth reward functions. Although PAO-DP does not always achieve the highest scores in this domain, its performance remains competitive and reliable. For the \textit{Med-Expert} dataset in HalfCheetah, PAO-DP scores 91.0, just shy of CQL's 91.6. In the \textit{Med-Replay} dataset, PAO-DP also performs well, particularly in \textit{Walker} with a score of 83.9, second only to IQL+EDP. The average performance in the Locomotion domain for PAO-DP is 78.5, showcasing its consistent effectiveness across various standard tasks.
While the Locomotion domain's smooth reward functions contrast with the sparse rewards found in domains like Kitchen, PAO-DP's ability to perform well across these diverse environments underscores its robustness and adaptability. This versatility highlights PAO-DP's effectiveness in handling a wide range of task complexities.

\textbf{Results for Peak Performance Evaluation: } 
To evaluate the peak performance potential of PAO-DP across different tasks, we conduct experiments using the OMS metric. As illustrated in Table \ref{oms_table_compare}, our method demonstrates competitive performance compared to other methods. In the \textit{FrankaKitchen} and \textit{AntMaze} domains, PAO-DP surpasses all baselines, achieving average scores of 80.6 and 96.7, respectively. This highlights its robustness in handling sparse rewards and complex trajectory tasks, showcasing its ability to optimize long-term value in challenging environments.
For the Adroit domain, PAO-DP scores 121.6, performing competitively against IQL+EDP (134.3) and CRR+EDP (116.9). 
The relatively smooth reward functions in the Locomotion domain contrast with the sparse rewards found in other domains, potentially diminishing PAO-DP's relative advantage. Despite this, PAO-DP's performance remains competitive, underscoring its versatility and robustness across diverse environments.

\begin{table}[t]
\small
\centering
\caption{Performance of the baselines and our method evaluated using the OMS metric.}
\setlength\tabcolsep{1.5pt} 
%\resizebox{0.9\textwidth}{!}{
\begin{tabular}{@{}cccccc@{}}
\toprule
Dataset                 & DIFF-QL & SfBC & CRR+EDP & IQL+EDP & \multicolumn{1}{c}{PAO-DP (Ours)} \\ \midrule
Average (FrankaKitchen) & 69.0    & 57.1 & 70.6    & 79.2    & 80.6 $\pm$ 13.79                     \\
Average (AntMaze)       & 69.6    & 74.2 & 76.7    & 89.2    & 96.7 $\pm$ 4.71                      \\
Average (Adroit)        & 65.1    & --      & 116.9   & 134.3   & 121.6 $\pm$ 3.06                     \\
Average (Locomotion)    & 88.0    & 75.6 & 87.3    & 84.7    & 84.1 $\pm$ 23.40                     \\ \bottomrule
\end{tabular}
\label{oms_table_compare}
\vspace{-0.2cm}
\end{table}

\vspace{-0.2cm}
\subsection{Ablation Studies}
\begin{figure}[ht]
\vspace{-0.5cm}
\centering
\subfigure[{}]{
\begin{minipage}[t]{0.33\linewidth}
\centering
\includegraphics[width=1.0\linewidth]{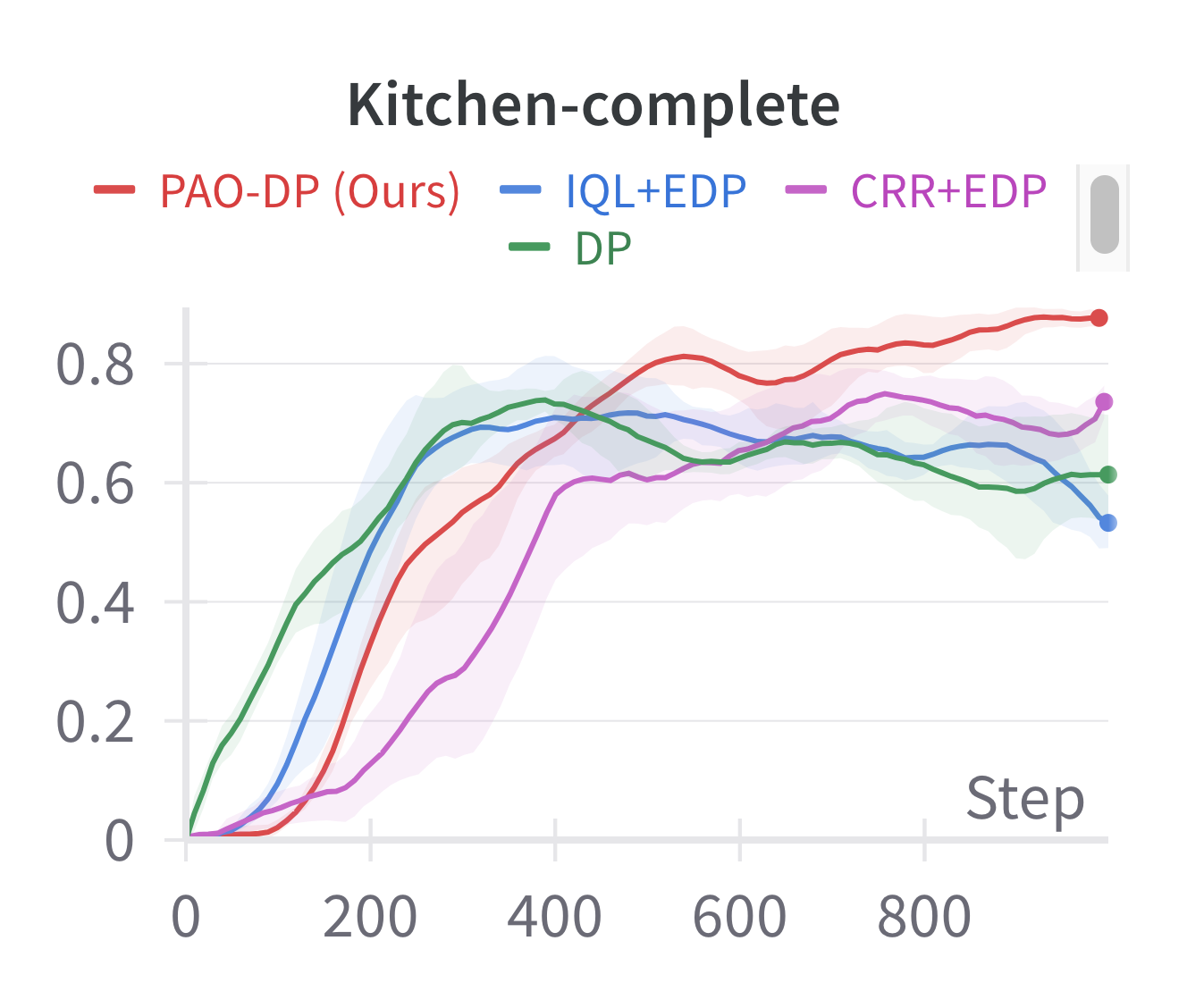}
\end{minipage}%
}%
\subfigure[{}]{
\begin{minipage}[t]{0.33\linewidth}
\centering
\includegraphics[width=1.0\linewidth]{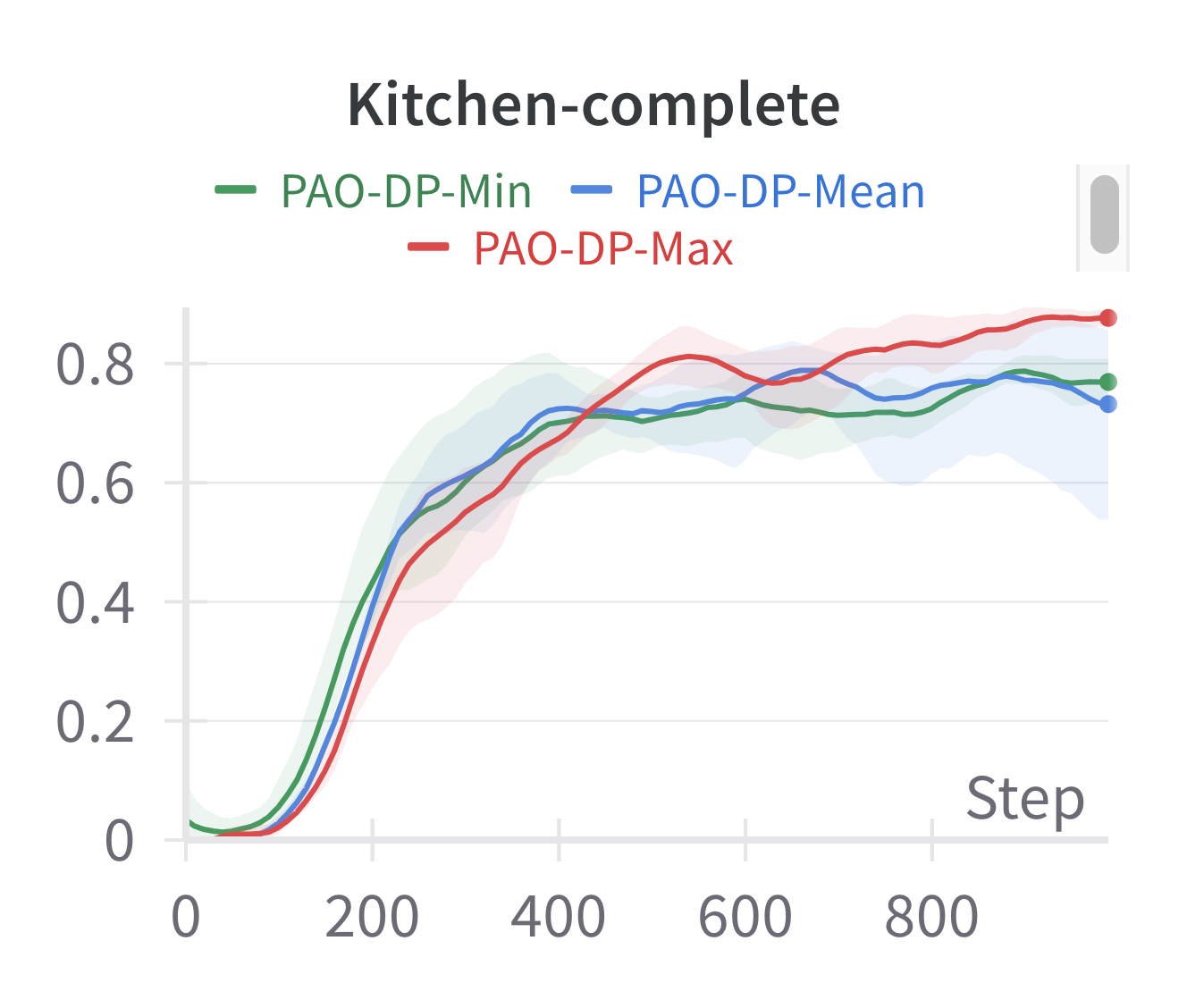}
\end{minipage}%
}%
\subfigure[{}]{
\begin{minipage}[t]{0.33\linewidth}
\centering
\includegraphics[width=1.0\linewidth]{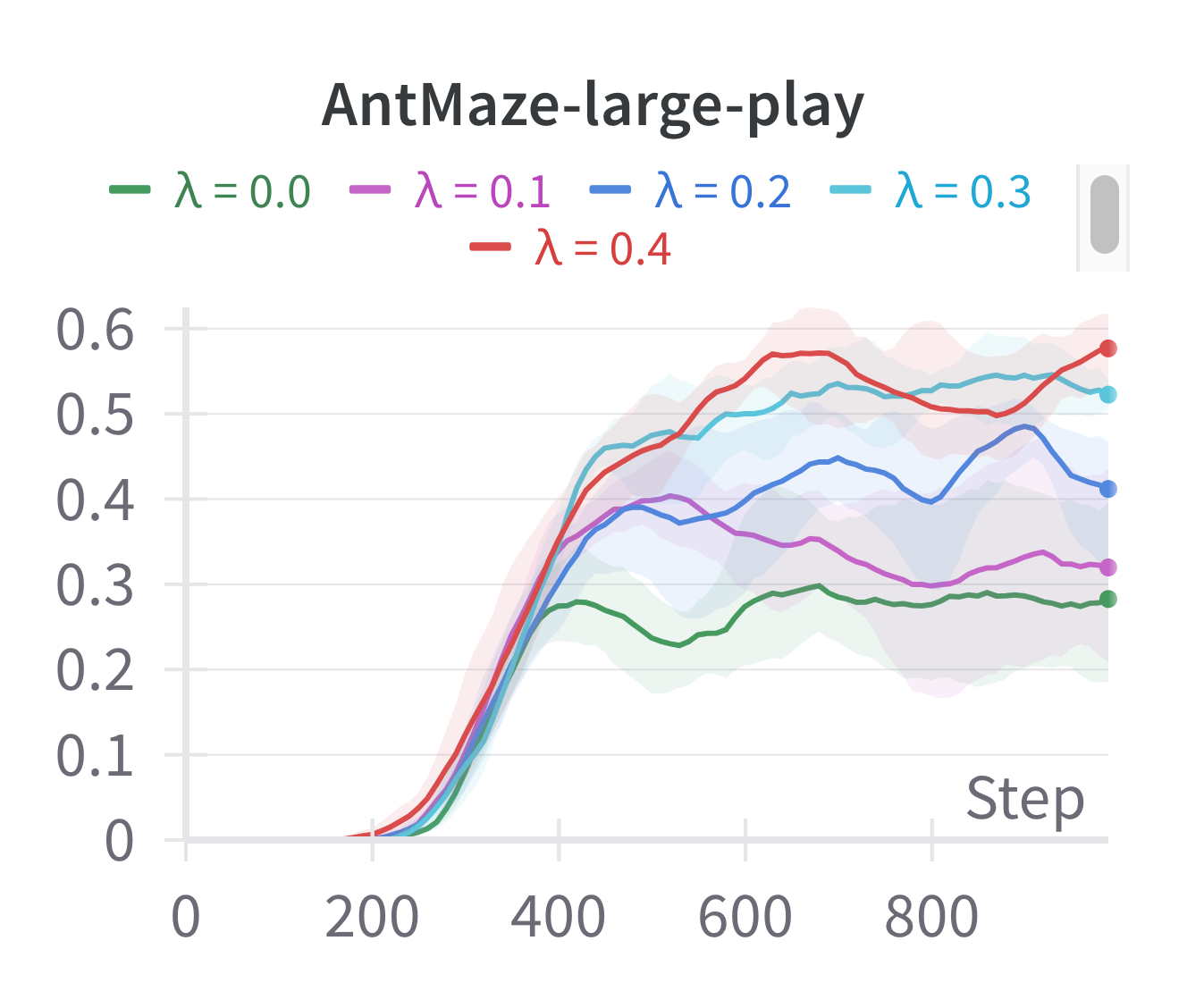}
\end{minipage}%
}%
\caption{ RAT curves of the ablation methods on representative tasks. The ordinate is the average normalized score.}
\label{fig:uav_training_resulsts_star}              
\centering
\vspace{-2mm}
\end{figure}

\textbf{Comparative Analysis of Diffusion Policy Methods: }
As illustrated in Figure \ref{fig:uav_training_resulsts_star}-(a), the ablation study on the \textit{Kitchen-complete} task underscores the effectiveness of the main component (i.e., preference optimization for policy improvement) of PAO-DP in comparison to other diffusion policy methods. PAO-DP significantly outperforms the DP method, which relies solely on policy cloning using a diffusion model, without any policy improvement. This highlights the benefits of policy improvement using preference optimization. Furthermore, PAO-DP demonstrates superior performance over IQL+EDP and CRR+EDP, both of which employ weighted regression for policy improvement. The sensitivity of WR to Q-value inaccuracies, coupled with constrained optimization using only sample actions within the fixed dataset, limits its efficacy. In contrast, PAO-DP's strategy of leveraging high-quality preferred actions and incorporating anti-noise preference optimization ensures more stable and robust training, leading to consistently better results in the later training.

\textbf{Analysis of Preferred-Action Sampling Strategies: }
Figure \ref{fig:uav_training_resulsts_star}-(b) shows the ablation study on the \textit{Kitchen-complete} task, comparing PAO-DP with different sampling strategies in preferred-action generation. PAO-DP-Max, PAO-DP-Min, and PAO-DP-mean respectively use the action with the highest, lowest, and middle weight \( \exp(\eta Q_\phi(s, a)) \) as the preferred action. PAO-DP-Max consistently achieves the best performance, compared with PAO-DP-Min and PAO-DP-Mean, which highlights the advantage of prioritizing actions with the highest weights. This study demonstrates that leveraging the important sampling technique for action selection leads to superior and more stable performance throughout training. More analysis results are shown in the Appendix \ref{Analysis of Preferred-Action Sampling Strategies}.

\textbf{Effect of Anti-Noise Optimization}
Figure \ref{fig:uav_training_resulsts_star}-(c) illustrates the ablation study on the \textit{AntMaze-large-play} task, examining the effectiveness of different values of $\lambda$ in our policy anti-noise preference optimization. As $\lambda$ increases from 0.0 to 0.4, we observe that the performance improves consistently throughout the training steps. The results demonstrate that incorporating a moderate level of noise helps stabilize the training process and enhances policy improvement, especially in low-quality datasets. More analysis results are shown in the Appendix \ref{The Impact of on Different Tasks}.
% This suggests that our anti-noise optimization approach, which calculates a cross-entropy loss with a small probability of label flipping, effectively mitigates the impact of Q-value estimation inaccuracies. 
% The impact of the noise parameter $\lambda$ on the policy performance is evaluated through an ablation study using the \textit{AntMaze-large-play} environment. The results, as depicted in the Figure \ref{fig:uav_training_resulsts_star}-(c), demonstrate the effect of varying $\lambda$ values on the learning step over time. 

% \begin{figure}[htbp]
% \centering
% \begin{subfigure}[b]{0.4\textwidth}
%     \centering
%     \includegraphics[width=\textwidth]{fig/ablation/amlp-lambda.svg}
%     \caption{subtitle}
%     \label{fig:sub1}
% \end{subfigure}
% \hfill
% \begin{subfigure}[b]{0.4\textwidth}
%     \centering
%     \includegraphics[width=\textwidth]{fig/ablation/amlp-lambda.svg}
%     \caption{subtitle}
%     \label{fig:sub2}
% \end{subfigure}
% \caption{title}
% \label{fig:test}
% \end{figure}

\subsection{Sensitivity Analysis to $\xi$}
\begin{figure}[ht]
\vspace{-0.6cm}
\centering
\subfigure[{\color{black}PAO-DP}]{
\begin{minipage}[t]{0.33\linewidth}
\centering
\includegraphics[width=1.0\linewidth]{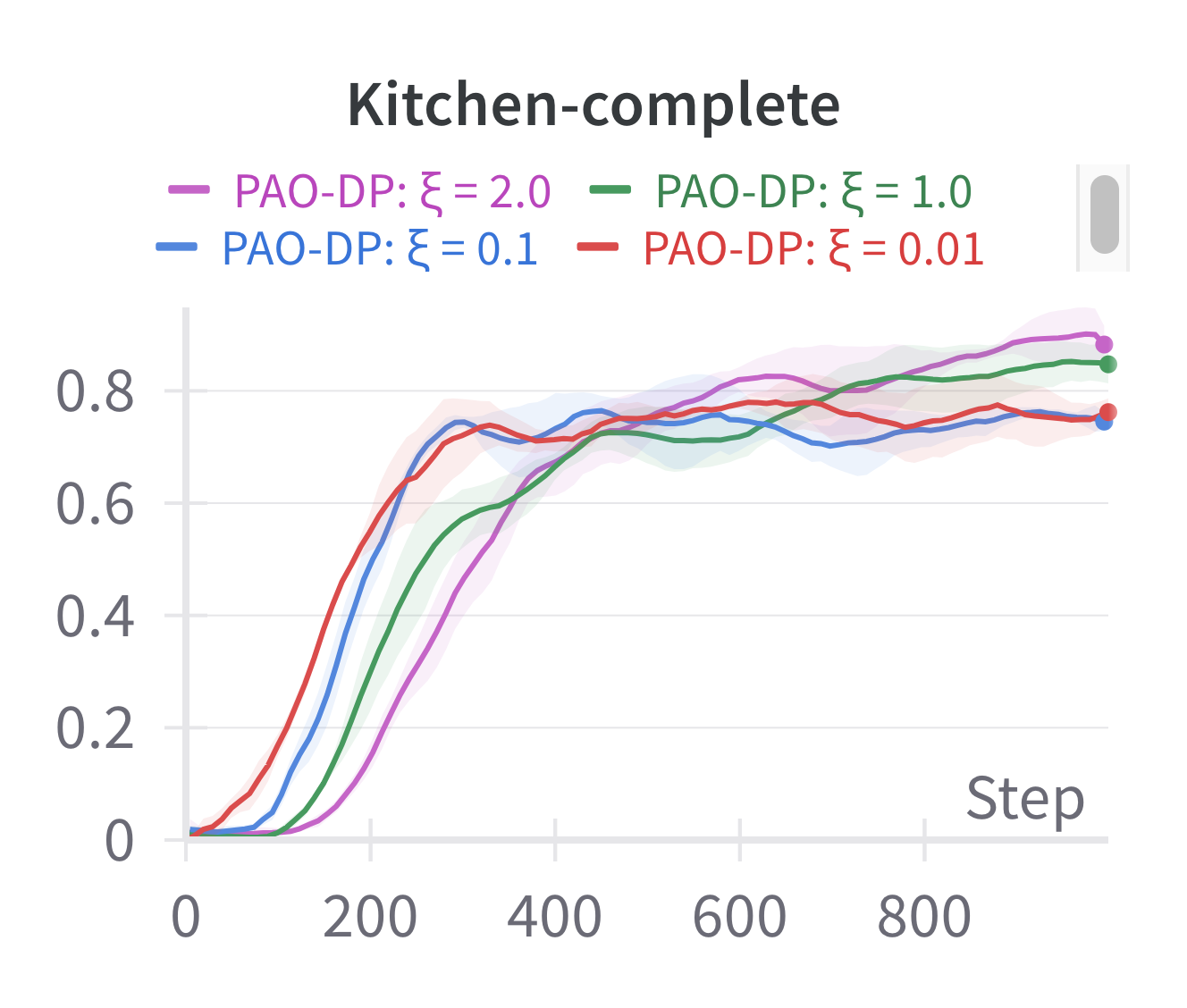}
\end{minipage}%
}%
\subfigure[{\color{black}IQL+EDP}]{
\begin{minipage}[t]{0.33\linewidth}
\centering
\includegraphics[width=1.0\linewidth]{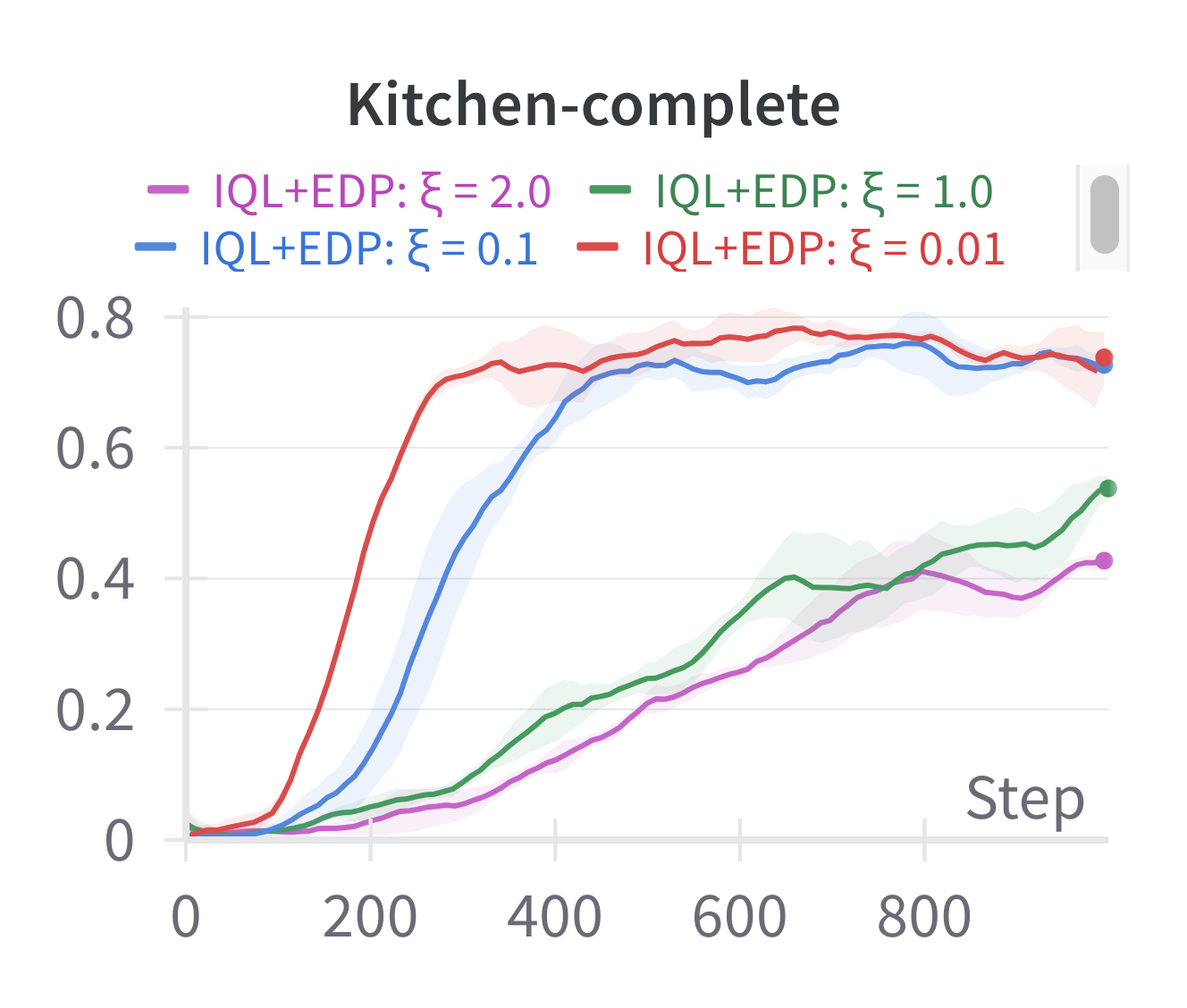}
\end{minipage}%
}%
\subfigure[{\color{black}CRR+EDP}]{
\begin{minipage}[t]{0.33\linewidth}
\centering
\includegraphics[width=1.0\linewidth]{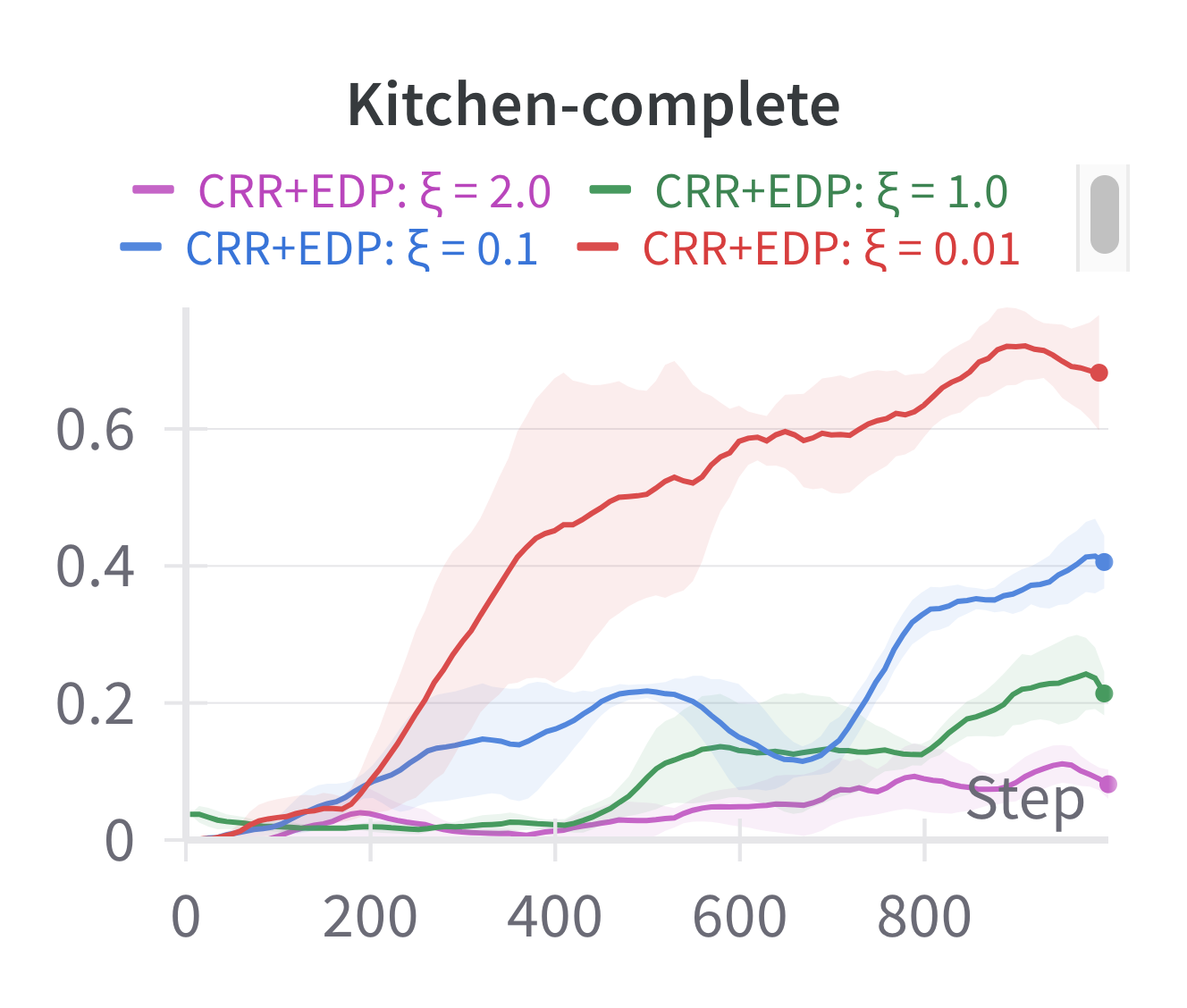}
\end{minipage}%
}%
\vspace{-0.1cm}
\caption{RAT curves for different methods with different $\xi$ on the Kitchen-complete task. The ordinate is the average normalized score.}
\label{fig:xi}              
\centering
\end{figure}

Figure \ref{fig:xi} illustrates the sensitivity analysis to the hyperparameter $\xi$ on the \textit{Kitchen-complete} task for PAO-DP, IQL+EDP, and CRR+EDP. The hyperparameter $\xi$ can reflect the effect of the policy improvement part in the entire training. In Figure \ref{fig:xi}-(a), PAO-DP demonstrates robust performance across different $\xi$ values, showing minimal sensitivity to changes in $\xi$. Conversely, 
Figures \ref{fig:xi}-(b) and \ref{fig:xi}-(c) illustrate that both IQL+EDP and CRR+EDP exhibit significant sensitivity to $\xi$, which indicates that the weighted regression for policy improvement has training instability. The analysis further verifies the robustness and effectiveness of the policy improvement part using the preference optimization of our method.

% The analysis indicates that PAO-DP maintains high performance regardless of the balance between policy cloning and improvement, showcasing its robustness and effectiveness. 

\vspace{-0.3cm}
\section{Related Work}
\vspace{-0.2cm}
% In this section, we extensively discuss the related work on offline RL. We primarily focus on two aspects: temporal-difference and sequence modeling methods. 
\textbf{Policy Regularization and Conservative Estimation}. Policy regularization and pessimistic conservative estimation are important approaches to addressing offline RL issues~\cite{an2021uncertainty,zhou2020plas,wei2021boosting,guan2023uac}. Policy regularization methods introduce regularization terms during the policy improvement process. For instance, the earliest proposed BCQ~\cite{fujimoto2019off} directly improves a policy perturbation model based on the behavior model of CVAE~\cite{sohn2015learning}. Based on this, some works such as BEAR~\cite{kumar2019stabilizing} and TD3+BC~\cite{fujimoto2021minimalist} incorporate the loss of behavior cloning into the policy improvement process by minimizing Maximum Mean Discrepancy~(MMD) or maximizing likelihood estimation. Additionally, BRAC~\cite{wu2019behavior} has explored regularization methods in KL divergence, MMD, and Wasserstein duality. On the other hand, some works~\cite{guan2023voce,anonymous2024poce} introduce pessimistic conservative estimation during policy evaluation to mitigate extrapolation errors caused by out-of-distribution actions. For instance, CQL~\cite{kumar2020conservative} supplements an additional penalty term in Q-values estimation to achieve a non-pointwise lower bound estimation of Q-values. Additionally, some works such as CRR~\cite{wang2020critic} and IQL~\cite{kostrikov2021offline} have achieved promising performance by leveraging advantage actions or within-distribution sampling.

\textbf{Diffusion and Transformer in RL.} The powerful sequence modeling capability and multi-modal representation ability of diffusion models and transformers enable them to be widely explored in RL~\cite{li2023hierarchical, ajay2022conditional, berrueta2024maximum, xu2022prompting}. Early proposals such as DT~\cite{chen2021decision} and TT~\cite{janner2021offline} are based on the transformer framework to generate state sequences or action-state sequences maximizing returns from expected trajectory rewards. On the other hand, the multi-modal representation capability of diffusion models enables them to exhibit significant advantages in offline RL tasks with datasets featuring multi-modal distributions~\cite{lu2023contrastive,hansen2023idql}.Diffusion-RL~\cite{wang2022diffusion} is the first exploration of diffusion policies in reinforcement learning policy modeling under multi-modal datasets. To guide policy sampling towards actions with high Q-values, some works~\cite{chen2023equidiff,kang2024efficient} construct policies as Q-value-guided conditional generative policies through weighted regression. In addition, some approaches~\cite{chen2022offline,lu2023contrastive} decouple policy learning into behavior learning and action evaluation and introduce in-distribution sampling to avoid extrapolation errors. To incorporate explicit Q-value guidance, some work~\cite{wang2022diffusion,ada2024diffusion} further subtracts the weighted expectation of Q-values during policy training to guide diffusion policy updates.

% Recently, diffusion models have been widely explored in various domains such as image generation, video generation, and behavior modeling~\cite{he2024diffusion} due to their strong sequential modeling and representation capabilities. some works have attempted to establish diffusion-based models~\cite{ajay2022conditional}, which maximize reward returns by sequentially modeling entire trajectories of states or action-state pairs. Based on this, some works~\cite{liang2023adaptdiffuser} have proposed decision-making models suitable for self-evolution by leveraging the data augmentation capability of diffusion models. Additionally, some works~\cite{wang2022diffusion} have employed the representational capacity of diffusion models to establish the policy models of RL. 

\vspace{-2.0mm}
\section{Conclusion}
\vspace{-1.0mm}
This paper proposes a novel preferred-action-optimized diffusion policy (PAO-DP) for offline RL. PAO-DP utilizes a preference model with preferred actions to improve the diffusion policy, instead of weighted regression. In particular, a conditional diffusion model is used to represent the distribution of behavior policy. Preferred actions are automatically generated through the critic function based on the diffusion model. Moreover, an anti-noise preference optimization using the preferred actions is designed to improve the policy. To the best of our knowledge, this paper is the first attempt to combine preference optimization and diffusion models for offline RL. Experimental results on the D4RL benchmark demonstrate that PAO-DP has competitive or superior performance compared to other RL methods. Ablation studies and parameter sensitivity analysis further validate the effectiveness of the key components of PAO-DP.

%\cite{10290933}

%\bibliographystyle{unsrtnat}
\bibliographystyle{unsrtnat}
\bibliography{references.bib}

% \usepackage{booktabs}
\clearpage
\appendix
\section{The overall design of PAO-DP}
\label{appendix:overall design}
\begin{figure*}[htp!] 
    \centering
    \includegraphics[width=1.0\textwidth]{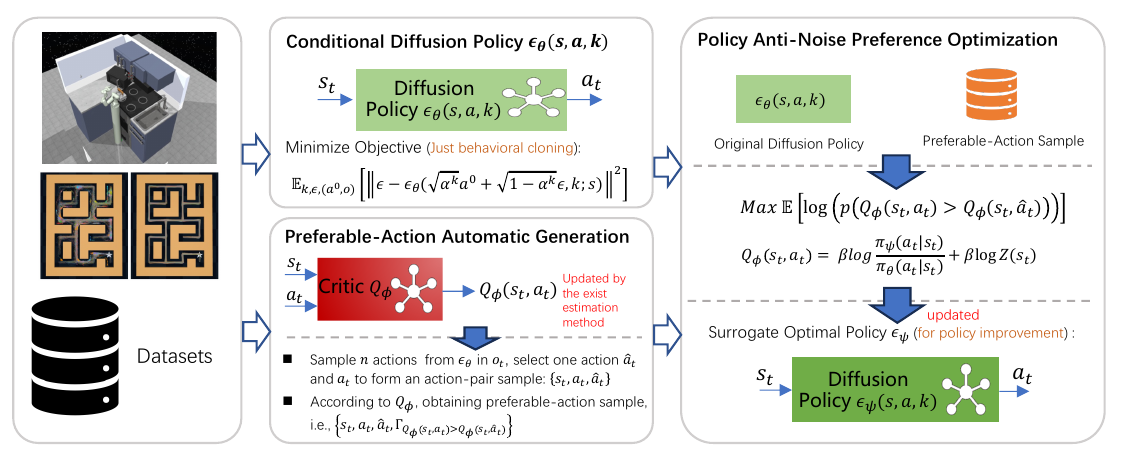}
    \caption{Overall design of the proposed method}
    \label{fig:overall_method}
\end{figure*}

To visually present the proposed method, the overall design of PAO-DP is further shown in Fig. \ref{fig:overall_method}.

\vspace{-0.3cm}

\section{Mathematical Derivations}
\subsection{The Optimal Closed-form Solution of the Regularized Objective in Offline RL}

In offline RL, we follow previous works~\cite{peng2019advantage, nair2020awac,chen2022offline} by maximizing the expected Q-value under policy $\pi(\bm{a}|\bm{s})$ while constraining the distance between policy $\pi(\bm{a}|\bm{s})$ and the behavior policy $\pi^{b}(\bm{a}|\bm{s})$. Therefore, the objective can be represented as $J(\pi)=\arg \max_{\pi} \mathbb{E}_{\bm{s}}\left[\sum_{\bm{a}}\pi(\bm{a}|\bm{s}) Q_{\phi}(\bm{s},\bm{a})\right] - {\eta}\mathbb{E}_{\bm{s}} \left[ D_{KL}(\pi(\cdot|\bm{s})||\pi^{b}(\cdot|\bm{s}))\right]$. Consequently, the optimal closed-form solution for this objective can be expressed as:
\small
\begin{equation}
    \begin{aligned}
    \label{appendix_b1_01} 
    \pi^{*}(\cdot|\bm{s}) =\frac{1}{Z(\bm{s})} \pi^{b}(\cdot|\bm{s}) \exp\left[\frac{1}{\eta} Q(\bm{s},\bm{a})\right].
    \end{aligned}
\end{equation}
\normalsize

\noindent \textbf{\textit{Proof.}} We review the objective of offline RL with a regularization term, and expanding the KL divergence term of the regularization yields the following expression:
\small
\begin{equation}
    \begin{aligned}
    \label{appendix_b1_1}
     & J(\pi)=\arg \max_{\pi} \mathbb{E}_{\bm{s}}\left[\sum_{\bm{a}}\pi(\bm{a}|\bm{s}) Q_{\phi}(\bm{s},\bm{a})\right] - {\eta}\mathbb{E}_{\bm{s}} \left[ D_{KL}(\pi(\cdot|\bm{s})||\pi^{b}(\cdot|\bm{s}))\right] \\
    &\quad \quad ~  =\arg \max_{\pi} \mathbb{E}_{\bm{s}}\left[\sum_{\bm{a}}\pi(\bm{a}|\bm{s}) Q_{\phi}(\bm{s},\bm{a})\right] - {\eta}\mathbb{E}_{\bm{s}}\left[\sum_{\bm{a}}\pi(\cdot|\bm{s})\log\frac{\pi(\cdot|\bm{s})}{\pi^{b}(\cdot|\bm{s})} \right] \\
    &\quad \quad ~  =\arg \max_{\pi} \mathbb{E}_{\bm{s}}\left[\sum_{\bm{a}}\pi(\bm{a}|\bm{s}) Q_{\phi}(\bm{s},\bm{a})\right] -{\eta}\mathbb{E}_{\bm{s}}\left[\sum_{\bm{a}}\pi(\cdot|\bm{s})\left[ \log \pi(\cdot|\bm{s})-\log\pi^{b}(\cdot|\bm{s})\right]\right].
    \end{aligned}
\end{equation}
\normalsize
To derive the closed-form solution for Eq.~(\ref{appendix_b1_1}), we differentiate Eq.~(\ref{appendix_b1_1}) concerning $\pi(\bm{a}|\bm{s})$ , resulting in the following expression:
\small
\begin{equation}
    \begin{aligned}
    \label{appendix_b1_2}
     & \mathbb{E}_{\bm{s}}\left[\sum_{\bm{a}}\pi(\bm{a}|\bm{s}) Q_{\phi}(\bm{s},\bm{a})\right] - {\eta}\mathbb{E}_{\bm{s}}\left[\sum_{\bm{a}}\pi(\cdot|\bm{s})\left[ \log \pi(\cdot|\bm{s})-\log\pi^{b}(\cdot|\bm{s})\right]\right]'_{{\pi}(\cdot|\bm{s})}\\
     &\overset{\{i\}}{=}\mathbb{E}_{\bm{s}}Q_{\phi}(\bm{s},\bm{a})-{\eta}\mathbb{E}_{\bm{s}}\left[\log \pi(\cdot|\bm{s})-\log \pi^{b}(\cdot|\bm{s}) +\sum_{\bm{a}} \pi (\cdot|\bm{s}) \frac{1}{\pi(\cdot|\bm{s})}\right]\\
     & =\mathbb{E}_{\bm{s}}\left[Q_{\phi}(\bm{s},\bm{a})-{\eta}\log\frac{\pi(\cdot|\bm{s})}{\pi^{b}(\cdot|\bm{s})} -{\eta}\right].
    \end{aligned}
\end{equation}
\normalsize
In practical computations, the KL divergence typically employs the natural logarithm base $e$. Consequently, in step $\{i\}$, $[\log\pi(\cdot|\bm{s})]'_{\pi(\cdot|\bm{s})}=\pi(\cdot|\bm{s})^{-1}$ holds true. To obtain the explicit expression of the closed-form solution $\pi^{*}(\cdot|s)$ for Eq.~(\ref{appendix_b1_1}), we set Eq~(\ref{appendix_b1_2}) to zero $\mathbb{E}_{\bm{s}}\left[Q_{\phi}(\bm{s},\bm{a})-{\eta}\log\frac{\pi(\cdot|\bm{s})}{\pi^{b}(\cdot|\bm{s})} -{\eta}\right]=0$, yielding the following expression:
\small
\begin{equation}
    \begin{aligned}
    \label{appendix_b1_3} 
     & Q_{\phi}(\bm{s},\bm{a})-{\eta}\log\frac{\pi^{*}(\cdot|\bm{s})}{\pi^{b}(\cdot|\bm{s})} -{\eta} = 0 
     \iff \log\frac{\pi^{*}(\cdot|\bm{s})}{\pi^{b}(\cdot|\bm{s})} = \eta Q(\bm{s},\bm{a}) -1 \\
     & \iff \pi^{*}(\cdot|\bm{s}) = \pi^{b}(\cdot|\bm{s}) \exp[\frac{1}{\eta} Q(\bm{s},\bm{a})-1] \\
     & \iff \pi^{*}(\cdot|\bm{s}) =\frac{1}{Z(\bm{s})} \pi^{b}(\cdot|\bm{s}) \exp[\frac{1}{\eta} Q(\bm{s},\bm{a})]
    \end{aligned}
\end{equation}
\normalsize
\noindent where $\pi^{*}$ is the optimal closed-form solution of Eq.~\ref{appendix_b1_1}, and ${Z}(\bm{s})=\exp(1)$ is the partition function. 

From Eq.~(\ref{appendix_b1_3}), it is evident that there is a direct relationship between the policy $\pi^{*}(\bm{a}|\bm{s})$ and the behavior policy $\pi^{b}(\bm{a}|\bm{s})$. However, accurately estimating the behavior policy $\pi^{b}(\bm{a}|\bm{s})$ is challenging. Therefore, following previous work\cite{peng2019advantage, wang2020critic, chen2020bail, chen2022offline}, we project the policy $\pi^{*}(\bm{a}|\bm{s})$ onto the parameterized policy $\pi_{\theta}(\bm{a}|\bm{s})$. Consequently, the final optimization objective can be expressed as:
\small
\begin{equation}\label{e2-11}
\begin{split} 
 &J(\pi_{\theta}) = \mathop{\arg\max}\limits_{\bm{\theta}} \; \mathbb{E}_{(\bm{s}, \bm{a}) \sim \mathcal{D}}\left[\frac{1}{Z(\bm{s})}\log \pi_{\theta}(\bm{a}|\bm{s})\exp(\frac{1}{\eta} Q_{\phi}(\bm{s},\bm{a}))\right].
\end{split}
\end{equation}
\normalsize
\noindent \textbf{\textit{Proof.}} We utilize the KL divergence to constrain the distance between $\pi^{*}$ and $\pi_{\theta}$, achieving the projection of $\pi^{*}$ onto the parameterized $\pi_{\theta}$. The 
The derivation process can be expressed as follows:
\small
\begin{equation}\label{appendix_b1_4}
    \begin{split} 
    & \mathop{\arg\min}\limits_{\theta} \; \mathbb{E}_{\bm{s} \sim \mathcal{D}}[D_{KL}(\pi^{*}(\cdot|\bm{s})||\pi_{\theta}(\cdot|\bm{s}))] \\
    & = \mathop{\arg\min}\limits_{\theta}\mathbb{E}_{\bm{s}\sim \mathcal{D}}\left[\sum_{\bm{a}} \pi^{*}(\cdot|\bm{s}) \left[ \log\pi^{*}(\cdot|\bm{s}) -\log\pi_{\theta}(\cdot|\bm{s})\right] \right]\\
    & = \mathop{\arg\min}\limits_{\theta}\mathbb{E}_{\bm{s}\sim \mathcal{D}}\mathbb{E}_{\bm{a}\sim {\pi^{*}(\cdot|\bm{s})}}\left[ \log \pi^{*}(\cdot|\bm{s})-\log \pi_{\theta}(\cdot|\bm{s})\right]\\
    & = \mathop{\arg\min}\limits_{\theta}\underbrace{\mathbb{E}_{\bm{s}\sim \mathcal{D}}\mathbb{E}_{\bm{a}\sim\pi^{*}(\cdot|\bm{s})}\log \pi^{*}(\cdot|\bm{s})}_{\mathcal{C}}-\mathbb{E}_{\bm{s}\sim \mathcal{D}}\mathbb{E}_{\bm{a}\sim\pi^{*}(\cdot|\bm{s})}\log \pi_{\theta}(\cdot|\bm{s})\\
    & \overset{\{i\}}{\iff}  \mathop{\arg\min}\limits_{\theta}-\mathbb{E}_{\bm{s}\sim \mathcal{D}}\mathbb{E}_{\bm{a}\sim\pi^{*}(\cdot|\bm{s})}\log \pi_{\theta}(\cdot|\bm{s})\\   
    & \iff  \mathop{\arg\max}\limits_{\theta}\mathbb{E}_{\bm{s}\sim \mathcal{D}}\mathbb{E}_{\bm{a}\sim\pi^{*}(\cdot|\bm{s})}\log \pi_{\theta}(\cdot|\bm{s})\\   
    & =  \mathop{\arg\max}\limits_{\theta}\mathbb{E}_{\bm{s}\sim \mathcal{D}}\left[\sum_{\bm{a}}\pi^{*}(\cdot|\bm{s})\log \pi_{\theta}(\cdot|\bm{s})\right]\\
    & =  \mathop{\arg\max}\limits_{\theta}\mathbb{E}_{\bm{s}\sim \mathcal{D}}\left[\sum_{\bm{a}}\pi^{b}(\cdot|\bm{s})\frac{\pi^{*}(\cdot|\bm{s})}{\pi^{b}(\cdot|\bm{s})}\log \pi_{\theta}(\cdot|\bm{s})\right] \\
    & =  \mathop{\arg\max}\limits_{\theta}\mathbb{E}_{\bm{s}\sim \mathcal{D}}\mathbb{E}_{\bm{a}\sim\pi^{b}(\cdot|\bm{s})} \left[\frac{\pi^{*}(\cdot|\bm{s})}{\pi^{b}(\cdot|\bm{s})}\log \pi_{\theta}(\cdot|\bm{s})\right]\\
    & \overset{\{ii\}}{=}  \mathop{\arg\max}\limits_{\theta}\mathbb{E}_{\bm{s}\sim \mathcal{D}}\mathbb{E}_{\bm{a}\sim\pi^{b}(\cdot|\bm{s})} \left[\frac{\frac{1}{Z(\bm{s})} \pi^{b}(\cdot|\bm{s}) \exp[\frac{1}{\eta }Q(\bm{s},\bm{a})]}{\pi^{b}(\cdot|\bm{s})}\log \pi_{\theta}(\cdot|\bm{s})\right]\\    
    & =  \mathop{\arg\max}\limits_{\theta}\mathbb{E}_{\bm{s}\sim \mathcal{D}}\mathbb{E}_{\bm{a}\sim\pi^{b}(\cdot|\bm{s})} \left[\frac{1}{Z(\bm{s})}\log \pi_{\theta}(\cdot|\bm{s})  \exp\left[\frac{1}{\eta} Q(\bm{s},\bm{a})\right]\right]\\
    & \overset{\{iii\}}{=}  \mathop{\arg\max}\limits_{\theta}\mathbb{E}_{\bm{s}\sim \mathcal{D}}\mathbb{E}_{\bm{a}\sim\mathcal{D}} \left[\frac{1}{Z(\bm{s})}\log \pi_{\theta}(\cdot|\bm{s})  \exp\left[\frac{1}{\eta} Q(\bm{s},\bm{a})\right]\right] \\
    & = \mathop{\arg\max}\limits_{\theta}\mathbb{E}_{(\bm{s},\bm{a})\sim \mathcal{D}} \left[\frac{1}{Z(\bm{s})}\log \pi_{\theta}(\cdot|\bm{s})  \exp\left[\frac{1}{\eta} Q(\bm{s},\bm{a})\right]\right]  
    \end{split}
\end{equation}
\normalsize
\noindent where ${Z}(\bm{s})$ is the same partition function as in Eq.~(\ref{appendix_b1_3}). In step $\{i\}$, since $\mathbb{E}_{\bm{s}\sim \mathcal{D}} \mathbb{E}_{\bm{a}\sim {\pi^{*}(\cdot|\bm{s})}}\left[ \log \pi^{*}(\cdot|\bm{s})\right]$ is a constant concerning the parameter $\theta$ during the optimization of Eq.~(\ref{appendix_b1_4}), it does not affect the maximization objective and can thus be directly ignored. In step $\{ii\}$, Eq.~(\ref{appendix_b1_3}) is substituted into the current equation. In step $\{iii\}$, $\pi^{b}(\bm{a}|\bm{s})$ is the behavior policy of the dataset $\mathcal{D}$, thus $\pi^{b}(\bm{a}|\bm{s})\sim\mathcal{D}$.

\subsection{Deriving the Preference Optimization Objective of PAO-DP Under the Bradley-Terry Model}
\label{Deriving the Preference Optimization Objective of PAO-DP}

In Section \ref{Policy Anti-Noise Preference Optimization}, the popular Bradley-Terry (BT) model \cite{bradley1952rank, rafailov2024direct}  is used to model the preference distribution $p^{*}$ for actions:
\small
\begin{equation} 
    \label{appendix_b2_1}
    p^{*}(\hat{\bm{a}}_{t} \succ \bm{a}_{t} | \bm{s}_{t}) = \frac{\exp (Q(\bm{s}_{t},\hat{\bm{a}}_{t}))}{\exp (Q(\bm{s}_{t},\hat{\bm{a}}_{t})) + \exp (Q(\bm{s}_{t},\bm{a}_{t}))}.\\
\end{equation}
\normalsize
Given Eq.~(\ref{appendix_b2_1}), we can directly derive the preference optimization objective of PAO-DP under the BT preference model. Meanwhile, in Section \ref{Policy Anti-Noise Preference Optimization}, based on Eq.~(\ref{e_2}), we can express the ground-truth Q-value through its corresponding optimal policy:
\small
\begin{equation} 
    \label{appendix_b2_2}
    Q(\bm{s}_{t},\bm{a}_{t})) = \eta \log \frac{\pi^{*}(\bm{a}_{t}|\bm{s}_{t})}{\pi^{b}(\bm{a}_{t}|\bm{s}_{t})}+\eta \log Z(\bm{s}),\\
\end{equation}
\normalsize
Substituting Eq.~(\ref{appendix_b2_2}) into Eq.~(\ref{appendix_b2_1}) we obtain:
% \small
% \begin{equation}
% \begin{split} 
%     \label{e_po_31}
% p^{*}(\hat{\bm{a}}_{t} \succ \bm{a}_{t} | \bm{s}_{t}) & = \frac{\exp \left(\eta \log \frac{\pi^{*}(\hat{\bm{a}}_{t}|\bm{s}_{t})}{\pi^{b}(\hat{\bm{a}}_{t}|\bm{s}_{t})}+\eta \log Z(s)\right)}{\exp \left(\eta \log \frac{\pi^{*}(\hat{\bm{a}}_{t}|\bm{s}_{t})}{\pi^{b}(\hat{\bm{a}}_{t}|\bm{s}_{t})}+\eta \log Z(s)\right) + \exp \left(\eta \log \frac{\pi^{*}(\bm{a}_{t}|\bm{s}_{t})}{\pi^{b}(\bm{a}_{t}|\bm{s}_{t})}+\eta \log Z(s)\right)} \\
% & = \frac{1}{1+\exp \left(\eta \log \frac{\pi^{*}(\bm{a}_{t}|\bm{s}_{t})}{\pi^{b}(\bm{a}_{t}|\bm{s}_{t})}-\eta \log \frac{\pi^{*}(\hat{\bm{a}}_{t}|\bm{s}_{t})}{\pi^{b}(\hat{\bm{a}}_{t}|\bm{s}_{t})}\right)} \\
% & = \sigma \left(\eta \log \frac{\pi^{*}(\hat{\bm{a}}_{t}|\bm{s}_{t})}{\pi^{b}(\hat{\bm{a}}_{t}|\bm{s}_{t})}-\eta \log \frac{\pi^{*}(\bm{a}_{t}|\bm{s}_{t})}{\pi^{b}(\bm{a}_{t}|\bm{s}_{t})}\right)
% \end{split}
% \end{equation}
% \normalsize
\small
\begin{equation}
\begin{split} 
    \label{e_po_31}
p^{*}(\hat{\bm{a}}_{t} \succ \bm{a}_{t} | \bm{s}_{t}) & = \frac{\exp \left(\eta \log \frac{\pi^{*}(\hat{\bm{a}}_{t}|\bm{s}_{t})}{\pi^{b}(\hat{\bm{a}}_{t}|\bm{s}_{t})}+\eta \log Z(s)\right)}{\exp \left(\eta \log \frac{\pi^{*}(\hat{\bm{a}}_{t}|\bm{s}_{t})}{\pi^{b}(\hat{\bm{a}}_{t}|\bm{s}_{t})}+\eta \log Z(s)\right) + \exp \left(\eta \log \frac{\pi^{*}(\bm{a}_{t}|\bm{s}_{t})}{\pi^{b}(\bm{a}_{t}|\bm{s}_{t})}+\eta \log Z(s)\right)} \\
& = \frac{\exp(\eta \log Z(\bm{s}))\exp \left(\eta \log \frac{\pi^{*}(\hat{\bm{a}}_{t}|\bm{s}_{t})}{\pi^{b}(\hat{\bm{a}}_{t}|\bm{s}_{t})}\right)}{\exp(\eta \log Z(\bm{s}))\left[\exp \left(\eta \log \frac{\pi^{*}(\hat{\bm{a}}_{t}|\bm{s}_{t})}{\pi^{b}(\hat{\bm{a}}_{t}|\bm{s}_{t})}\right) + \exp \left(\eta \log \frac{\pi^{*}(\bm{a}_{t}|\bm{s}_{t})}{\pi^{b}(\bm{a}_{t}|\bm{s}_{t})}\right)\right]} \\
& = \frac{\exp \left(\eta \log \frac{\pi^{*}(\hat{\bm{a}}_{t}|\bm{s}_{t})}{\pi^{b}(\hat{\bm{a}}_{t}|\bm{s}_{t})}\right)}{\exp \left(\eta \log \frac{\pi^{*}(\hat{\bm{a}}_{t}|\bm{s}_{t})}{\pi^{b}(\hat{\bm{a}}_{t}|\bm{s}_{t})}\right) + \exp \left(\eta \log \frac{\pi^{*}(\bm{a}_{t}|\bm{s}_{t})}{\pi^{b}(\bm{a}_{t}|\bm{s}_{t})}\right)} \\
& = \frac{1}{1+\exp \left(\eta \log \frac{\pi^{*}(\bm{a}_{t}|\bm{s}_{t})}{\pi^{b}(\bm{a}_{t}|\bm{s}_{t})}-\eta \log \frac{\pi^{*}(\hat{\bm{a}}_{t}|\bm{s}_{t})}{\pi^{b}(\hat{\bm{a}}_{t}|\bm{s}_{t})}\right)} \\
& = \sigma \left(\eta \log \frac{\pi^{*}(\hat{\bm{a}}_{t}|\bm{s}_{t})}{\pi^{b}(\hat{\bm{a}}_{t}|\bm{s}_{t})}-\eta \log \frac{\pi^{*}(\bm{a}_{t}|\bm{s}_{t})}{\pi^{b}(\bm{a}_{t}|\bm{s}_{t})}\right),
\end{split}
\end{equation}
\normalsize
$\sigma$ is the logistic function. The final line represents the per-instance loss in Eq.~(\ref{e_po_4}).

\section{More Results}
\label{appendix:More Results}

\begin{table}[htp!]
\centering
\small
\caption{Performance of the baselines and PAO-OP evaluated using the OMS metric.}
\begin{tabular}{@{}cccccc@{}}
\toprule
\multicolumn{1}{c}{Dataset} & \multicolumn{1}{c}{Environment} & DIFF-QL & CRR+EDP & IQL+EDP & \multicolumn{1}{c}{\textbf{PAO-DP (Ours)}} \\ \midrule
Complete            & Kitchen            & 84.0  & 95.8  & 95.0  & 100.0 $\pm$ 0.00 \\
Partial             & Kitchen            & 60.5  & 56.7  & 72.5  & 70.0 $\pm$ 2.04  \\
Mixed               & Kitchen            & 62.5  & 59.2  & 70.0  & 71.8 $\pm$ 0.94  \\
\multicolumn{2}{c}{Average   (FrankaKitchen)}                 & 69.0    & 70.6    & 79.2    & 80.6 $\pm$ 13.79                     \\
\hline
Default             & AntMaze-umaze      & 93.6  & 98.0  & 98.0  & 100.0 $\pm$ 0.0  \\
Diverse             & AntMaze-umaze      & 66.2  & 80.0  & 90.0  & 100.0 $\pm$ 0.0  \\
Play                & AntMaze-medium     & 76.6  & 82.0  & 89.0  & 100.0 $\pm$ 0.0  \\
Diverse             & AntMaze-medium     & 78.6  & 72.0  & 88.0  & 100.0 $\pm$ 0.0  \\
Play                & AntMaze-large      & 46.4  & 57.0  & 52.0  & 90.0 $\pm$ 0.0   \\
Diverse             & AntMaze-large      & 56.6  & 71.0  & 68.0  & 90.0 $\pm$ 0.0   \\
\multicolumn{2}{c}{Average (AntMaze)}    & 69.6  & 76.7  & 89.2  & 96.7 $\pm$ 4.71  \\
\hline
human               & pen                & 72.8  & 127.8 & 130.3 & 124.1 $\pm$ 1.59 \\
cloned              & pen                & 57.3  & 106.0 & 138.2 & 119.2 $\pm$ 2.05 \\
\multicolumn{2}{c}{Average (Adroit)}     & 65.1  & 116.9 & 134.3 & 121.6 $\pm$ 3.06 \\
\hline
Medium-Expert       & HalfCheetah        & 96.8  & 93.5  & 80.9  & 92.4 $\pm$ 0.27  \\
Medium-Expert       & Hopper             & 111.1 & 109.4 & 95.7  & 111.1 $\pm$ 0.97 \\
Medium-Expert       & Walker             & 110.1 & 112.3 & 111.5 & 110.6 $\pm$ 0.49 \\
Medium              & HalfCheetah        & 51.1  & 50.2  & 48.7  & 47.4 $\pm$ 0.25  \\
Medium              & Hopper             & 90.5  & 95.0  & 97.3  & 71.4 $\pm$ 1.27  \\
Medium              & Walker             & 87.0  & 85.8  & 88.7  & 88.0 $\pm$ 0.31  \\
Medium-Replay       & HalfCheetah        & 47.8  & 47.8  & 45.5  & 44.5 $\pm$ 0.10  \\
Medium-Replay       & Hopper             & 101.3 & 101.7 & 100.9 & 100.8 $\pm$ 0.49 \\
Medium-Replay       & Walker             & 95.5  & 89.8  & 93.4  & 90.9 $\pm$ 0.67  \\
\multicolumn{2}{c}{Average (Locomotion)} & 88.0  & 87.3  & 84.7  & 84.1 $\pm$ 23.40 \\ \bottomrule
\end{tabular}
\label{tab:detail_oms_per}
\end{table}

Table \ref{tab:detail_oms_per} presents the detailed performance of various baseline methods and our proposed PAO-DP evaluated using the OMS metric across multiple environments. For the \textit{FrankaKitchen} domain, PAO-DP consistently outperforms other methods, particularly excelling in the \textit{Complete} dataset. This indicates PAO-DP's ability to handle tasks requiring sequential subtask completion.
In the \textit{AntMaze} domain, PAO-DP again demonstrates superior performance, achieving perfect scores in different datasets. This highlights the method's effectiveness in navigating complex mazes and dealing with sparse rewards.
For the \textit{Adroit} domain, although PAO-DP's average score of 121.6 is slightly lower than IQL+EDP, it remains competitive, showing strong performance in human and cloned tasks. This reflects its capability in high-dimensional robotic manipulation tasks.
In the \textit{Locomotion} domain, the performance is more varied. PAO-DP performs well in some tasks but shows a wider range in scores. This variability suggests that while PAO-DP is versatile, its relative advantage may be reduced in environments with smoother reward functions compared to sparse-reward settings. This may be because in sparse-reward environments, preferred actions with significant quality differences can be generated, efficiently promoting preference optimization for policy improvement, but less in smooth-reward environments. Overall, PAO-DP demonstrates robust and competitive performance across diverse and challenging environments, particularly excelling in scenarios with complex trajectories and sparse rewards.

\section{More Ablation Analysis}

\begin{figure}[htbp]
\centering
\subfigure[{\color{black}$\lambda = 0.0 $}]{
\begin{minipage}[t]{0.33\linewidth}
\centering
\includegraphics[width=1.0\linewidth]{fig/ablation-1/kc-sample-0.0-1.png}
\end{minipage}%
}%
\subfigure[{\color{black}$\lambda = 0.0 $}]{
\begin{minipage}[t]{0.33\linewidth}
\centering
\includegraphics[width=1.0\linewidth]{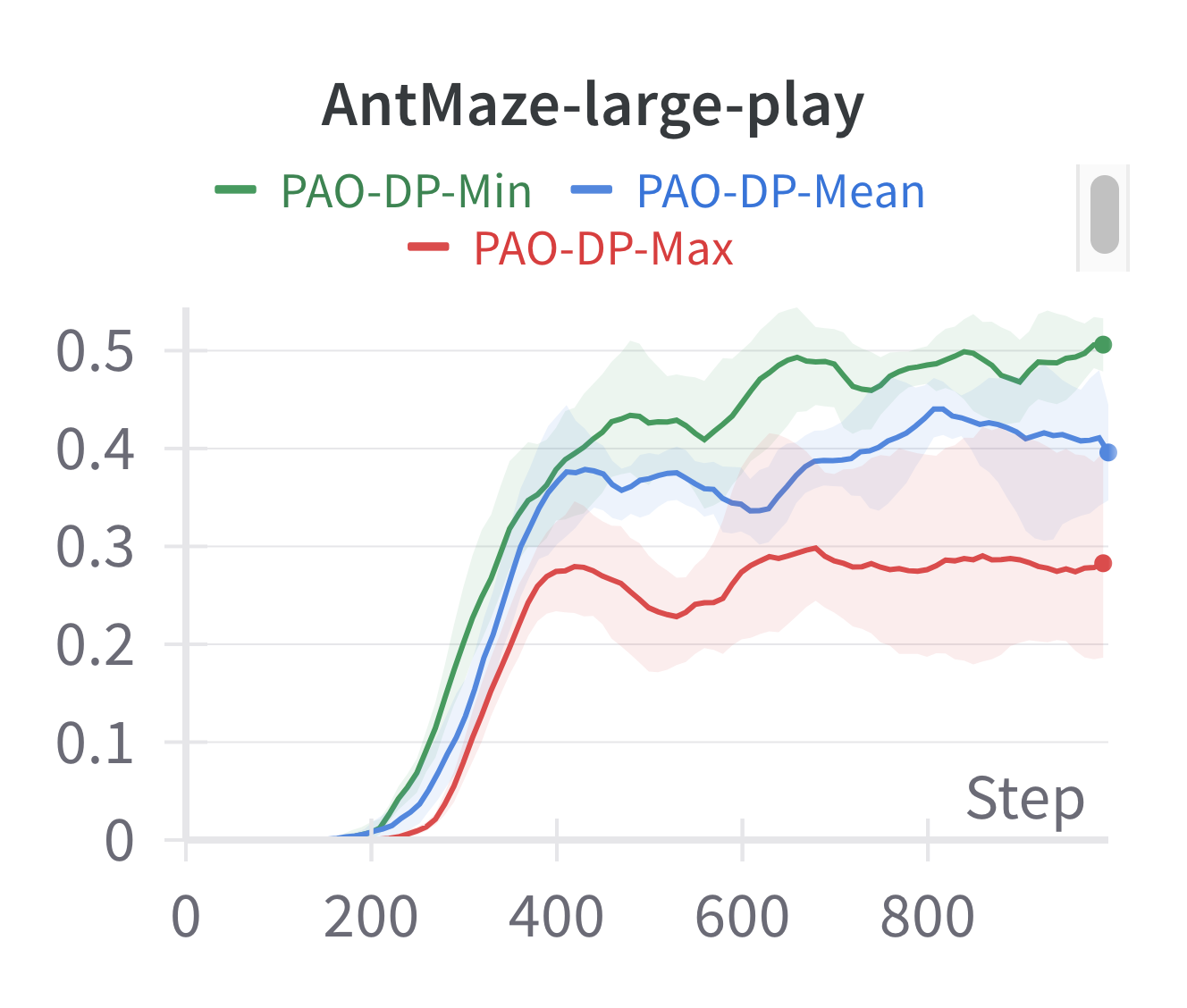}
\end{minipage}%
}%
\subfigure[{\color{black}$\lambda = 0.4 $}]{
\begin{minipage}[t]{0.33\linewidth}
\centering
\includegraphics[width=1.0\linewidth]{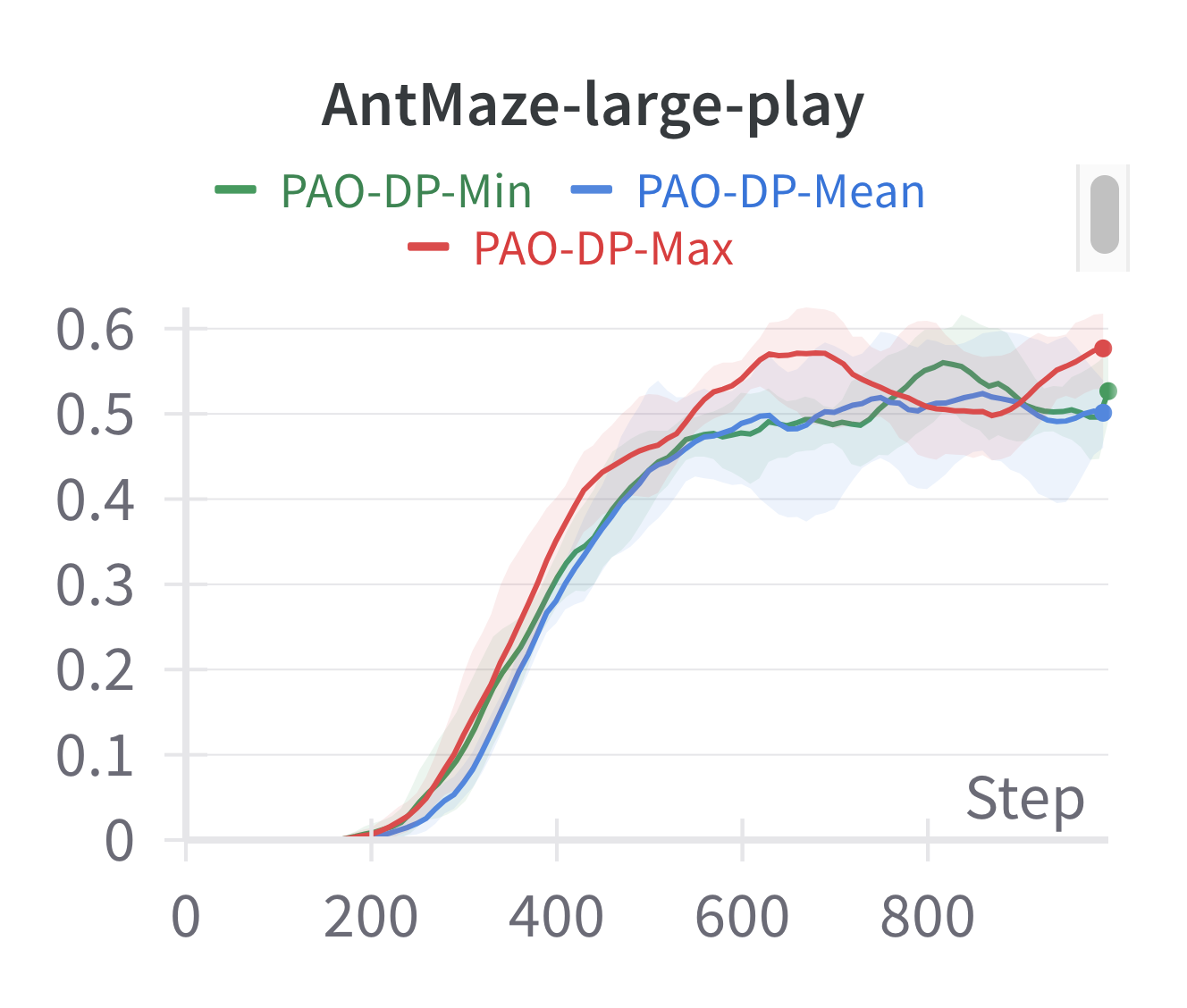}
\end{minipage}%
}%
\caption{Performance of PAO-DP variants in \textit{Kitchen-complete} and \textit{AntMaze-large-play} environments, comparing different sampling strategies with $\lambda = 0.0$ and $\lambda = 0.4$.}
\label{fig:ablation_study_1}              
\centering
\end{figure}

\textbf{Analysis of Preferred-Action Sampling Strategies. }
\label{Analysis of Preferred-Action Sampling Strategies}
% We investigates the impact of different sampling strategies and the proposed anti-noise optimization method in the PAO-DP framework. 
Figures \ref{fig:ablation_study_1}-(a) and \ref{fig:ablation_study_1}-(b) explore scenarios where $\lambda = 0.0$, indicating the absence of the anti-noise strategy. In the relatively noise-free \textit{Kitchen-complete} dataset, PAO-DP-Max shows superior performance, suggesting that in datasets with minimal noise, the highest sampling weight is reliable and effective because high-quality actions may be close to the distribution of optimal actions. Conversely, in the more complex and randomization \textit{AntMaze-large-play} dataset, PAO-DP-Max performs less reliably, highlighting that in challenging settings, the highest sampling weight may lead to sub-optimal policy updates. Figures \ref{fig:ablation_study_1}-(b) and \ref{fig:ablation_study_1}-(c) further illustrate the impact of introducing the anti-noise mechanism by varying $\lambda$. With $\lambda = 0.0$, PAO-DP-Max's performance is significantly lower compared to PAO-DP-Mean and PAO-DP-Min, indicating unreliability in relatively noisy datasets. However, increasing $\lambda$ to 0.4 shows a recovery in PAO-DP-Max's performance, aligning it closely with optimal results observed in PAO-DP-Mean and PAO-DP-Min. This recovery underscores the effectiveness of the anti-noise strategy in mitigating the adverse effects of noise, enhancing robustness, and improving policy reliability. This analysis highlights the critical role of preferred-action sampling strategies for policy improvement and the need for rational selection of the sampling strategies based on the anti-noise parameter $\lambda$.

\begin{figure}[htbp]
\centering
\subfigure[{\color{black}}]{
\begin{minipage}[t]{0.5\linewidth}
\centering
\includegraphics[width=0.9\linewidth]{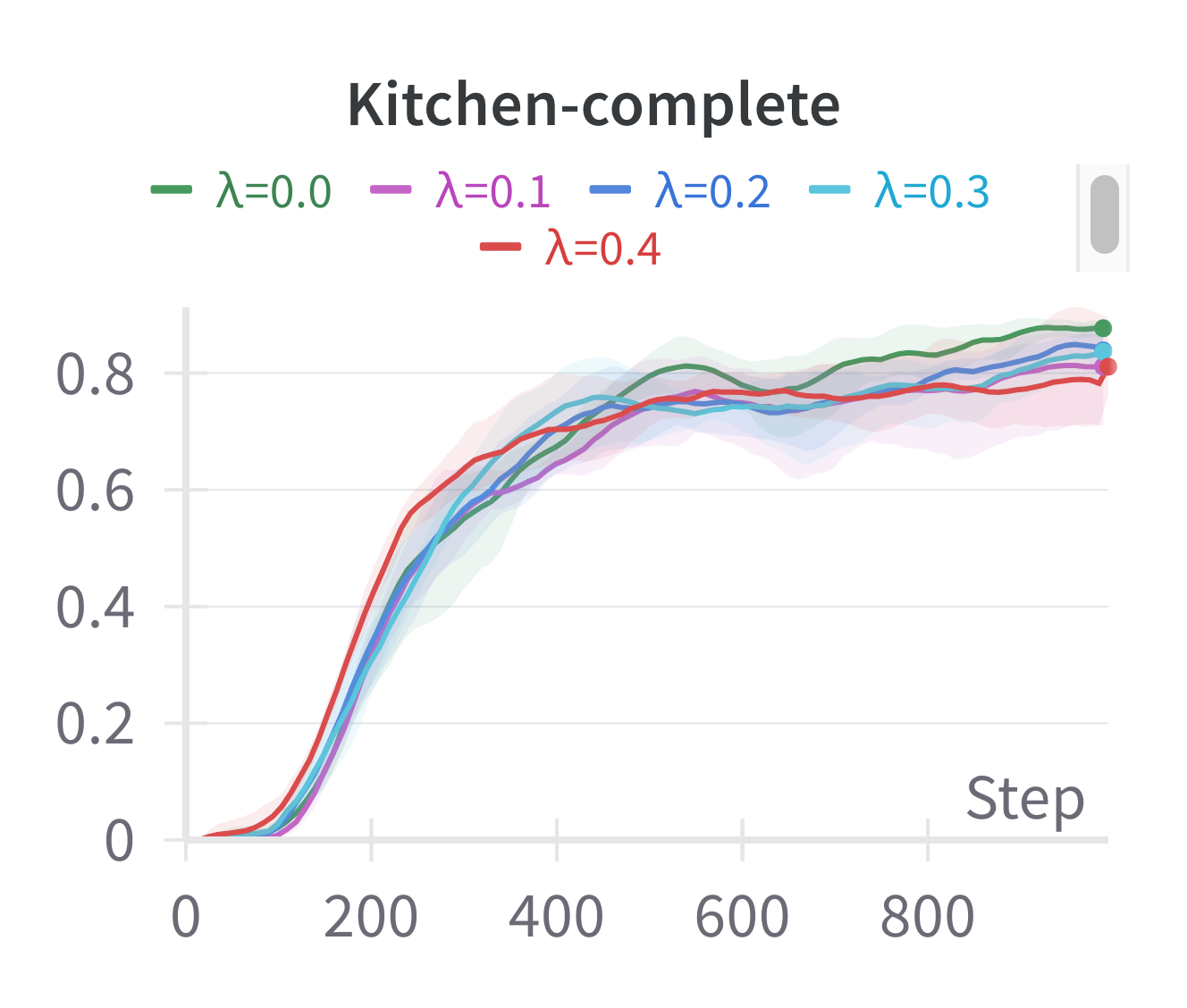}
\end{minipage}%
}%
\subfigure[{\color{black}}]{
\begin{minipage}[t]{0.5\linewidth}
\centering
\includegraphics[width=0.9\linewidth]{fig/ablation-1/amlp-lambda.png}
\end{minipage}%
}%
\caption{Impact of varying $\lambda$ values on PAO-DP performance in Kitchen-complete and AntMaze-large-play environments.}
\label{fig:ablation_study_lambda}              
\centering
\end{figure}

\textbf{The Impact of $\lambda$ on Different Tasks.}
\label{The Impact of on Different Tasks}
% The analysis investigates the impact of varying $\lambda$ values on the performance of the PAO-DP framework in two different environments: Kitchen-complete and AntMaze-large-play. 
Figure \ref{fig:ablation_study_lambda}-(a) illustrates that in the relatively noise-free \textit{Kitchen-complete} dataset, PAO-DP-Max with different $\lambda$ values yields similar performance outcomes, indicating the minimal impact of the anti-noise strategy. This suggests that in low-noise datasets, policy performance remains robust across various $\lambda$ settings, with the highest sampling weight (PAO-DP-Max) consistently achieving excellent results.
In contrast, Figure \ref{fig:ablation_study_lambda}-(b) reveals significant performance variations in the \textit{AntMaze-large-play} environment as $\lambda$ increases. With $\lambda = 0.0$, performance is notably lower, especially for PAO-DP-Max (See Figure \ref{fig:ablation_study_1}-(b)), highlighting the challenges posed by the complex and noisy nature of this environment. As $\lambda$ increases, performance improves, particularly for $\lambda = 0.4$, where PAO-DP-Max aligns closely with PAO-DP-Mean and PAO-DP-Min (See Figure \ref{fig:ablation_study_1}-(c)). This indicates that the proposed anti-noise mechanism effectively mitigates noise, enhancing policy reliability and robustness in challenging noisy, complex settings. 
 % Higher $\lambda$ improves robustness in noisy, complex scenarios.

\section{Hyperparameters}
\label{Hyperparameters}

The detailed hyperparameters of the proposed method are presented in Table~\ref{table:hyper}.

\begin{table}[htp!]
\caption{Hyperparameters of PAO-DP.}
\label{table:hyper} 
\begin{tabular}{@{}cccccccc@{}}
\toprule
\multicolumn{1}{c}{Dataset} & \multicolumn{1}{c}{Environment} & Learning rate & Epochs & Batch size & $\eta$ & $\lambda$ & $\xi$\\ \midrule
Complete            & Kitchen            & 3e-4  & 1000  & 256 & 0.1 & 0.0  & 1.0 \\
Partial             & Kitchen            & 3e-4  & 1000  & 256 & 0.1 & 0.2  & 1.0 \\
Mixed               & Kitchen            & 3e-4  & 1000  & 256 & 0.1 & 0.2  & 1.0 \\
\hline
Default             & AntMaze-umaze      & 3e-4  & 1000  & 256 & 0.1 & 0.2  & 1.0 \\
Diverse             & AntMaze-umaze      & 3e-4  & 1000  & 256 & 0.1 & 0.2  & 1.0 \\
Play                & AntMaze-medium     & 3e-4  & 1000  & 256 & 0.1 & 0.1  & 1.0 \\
Diverse             & AntMaze-medium     & 2e-4  & 1000  & 256 & 0.1 & 0.5  & 1.0 \\
Play                & AntMaze-large      & 1e-4  & 1000  & 256 & 0.1 & 0.4  & 1.0 \\
Diverse             & AntMaze-large      & 1e-4  & 1000  & 256 & 0.1 & 0.4  & 1.0 \\
\hline
human               & pen                & 1e-4  & 1000  & 256 & 0.1 & 0.4 & 1.0 \\
cloned              & pen                & 5e-5  & 1000  & 256 & 0.1 & 0.2  & 1.0\\
\hline
Medium-Expert       & HalfCheetah        & 3e-4  & 2000  & 256 & 0.1 & 0.2  & 1.0\\
Medium-Expert       & Hopper             & 3e-4  & 2000  & 256 & 0.1 & 0.2  & 1.0\\
Medium-Expert       & Walker             & 3e-4  & 2000  & 256 & 0.1 & 0.2  & 1.0\\
Medium              & HalfCheetah        & 3e-4  & 2000  & 256 & 0.1 & 0.2  & 1.0\\
Medium              & Hopper             & 3e-4  & 2000  & 256 & 0.1 & 0.2  & 1.0\\
Medium              & Walker             & 3e-4  & 2000  & 256 & 0.1 & 0.1  & 1.0\\
Medium-Replay       & HalfCheetah        & 3e-4  & 2000  & 256 & 0.1 & 0.1  & 1.0\\
Medium-Replay       & Hopper             & 3e-4  & 2000  & 256 & 0.1 & 0.1  & 1.0\\
Medium-Replay       & Walker             & 3e-4  & 2000  & 256 & 0.1 & 0.4  & 1.0\\\bottomrule
\end{tabular}
\end{table}

\section{Limitations and Future Work}
% 受Q值的影响，
% 轨迹->做偏好选择,预测未来. 
The effectiveness of the proposed PAO-DP hinges on accurate Q-value estimation to select preferred actions. However, offline RL continues to grapple with the out-of-distribution problem, where discrepancies between evaluation and training data may lead to Q-value estimation errors. These inaccuracies propagate throughout the learning process, resulting in suboptimal policy performance. Additionally, the accuracy of Q-values is impacted by the 
 quality and complexity of datasets, further challenging the robustness of the policy. 
 
Future research could enhance PAO-DP by incorporating trajectory-based preference optimization. We directly generate preferred trajectory data for preference optimization, instead of preferred action data using Q-values. This avoids the OOD problem using Q-values. We can compare two trajectories based on success or time to generate preferred trajectory data. This approach is expected to improve policy robustness and generalization, especially in complex environments. In addition, developing methods to encode and utilize trajectory information, possibly using advanced sequence modeling techniques, holds promise for more stable and strong offline RL.

% Future research could enhance PAO-DP by incorporating trajectory-based preference learning. Trajectories offer richer contextual information by capturing cumulative effects and temporal dependencies, which can potentially reduce Q-value estimation errors. This approach is expected to improve policy robustness and generalization, especially in complex environments. Developing methods to encode and utilize trajectory information, possibly using advanced sequence modeling techniques, holds promise for more stable offline RL.

%%%%%%%%%%%%%%%%%%%%%%%%%%%%%%%%%%%%%%%%%%%%%%%%%%%%%%%%%%%%

\end{document}